\documentclass{article}
\usepackage[margin=2cm]{geometry}
\usepackage{authblk}
\usepackage{enumitem}
\setlist{nosep} 

\usepackage{lmodern}

\usepackage{amsmath,amssymb,amsthm,bm,dsfont}
\usepackage{float}
\usepackage{xcolor}
\usepackage{mathtools}
\usepackage{booktabs,multirow,bigstrut,makecell}
\usepackage{algorithm}
\usepackage{algorithmic}
\usepackage{comment}
\usepackage{tikz}
\usetikzlibrary{positioning,fit,quotes,arrows,calc}
\usepackage{caption}
\usepackage{subcaption}
\usepackage{multicol}

\newcommand{\sam}[1]{
  {\color{blue} [SAM: {#1}]}
}

\definecolor{dkgreen}{rgb}{0,0.6,0}
\definecolor{gray}{rgb}{0.5,0.5,0.5}
\definecolor{mauve}{rgb}{0.58,0,0.82}

\definecolor{color1}{HTML}{53446B}
\definecolor{color2}{HTML}{807DBA}
\definecolor{color3}{HTML}{C0AECE}
\definecolor{color4}{HTML}{CE9292}
\definecolor{color5}{HTML}{DAB6B6}
\definecolor{color6}{HTML}{A40407}
\definecolor{color7}{HTML}{7484A8}
\definecolor{color8}{HTML}{C3965E}
\definecolor{color9}{HTML}{7279A4}

\newtheorem{thm}{Theorem}
\newtheorem{lem}[thm]{Lemma}
\newtheorem{prop}[thm]{Proposition}

\theoremstyle{definition}
\newtheorem{dfn}{Definition}

\theoremstyle{remark}
\newtheorem*{rmk}{Remark}

\newcommand{\Ib}{\mathbf{I}}

\newcommand{\Lb}{\mathbf{L}}

\newcommand{\Xb}{\mathbf{X}}
\newcommand{\Yb}{\mathbf{Y}}

\newcommand{\bX}{\bm{X}}

\newcommand{\bbeta}{\bm{\beta}}

\newcommand{\bmu}{\bm{\mu}}

\newcommand{\btheta}{\bm{\theta}}

\newcommand{\bpi}{\bm{\pi}}
\newcommand{\bnu}{\bm{\nu}}

\newcommand{\bLambda}{\bm{\Lambda}}
\newcommand{\bSigma}{\bm{\Sigma}}

\newcommand{\bbE}{\mathbb{E}}

\newcommand{\bbO}{\mathbb{O}}
\newcommand{\bbP}{\mathbb{P}}

\newcommand{\bbR}{\mathbb{R}}

\newcommand{\bbT}{\mathbb{T}}

\newcommand{\cA}{\mathcal{A}}
\newcommand{\cB}{\mathcal{B}}

\newcommand{\cI}{\mathcal{I}}

\newcommand{\cO}{\mathcal{O}}

\newcommand{\cQ}{\mathcal{Q}}

\newcommand{\cS}{\mathcal{S}}
\newcommand{\cT}{\mathcal{T}}
\newcommand{\cU}{\mathcal{U}}
\newcommand{\cV}{\mathcal{V}}

\newcommand{\cZ}{\mathcal{Z}}

\newcommand{\argmin}{\mathop{\mathrm{argmin}}}
\newcommand{\argmax}{\mathop{\mathrm{argmax}}}

\newcommand{\bzero}{{\mathbf{0}}}   
\newcommand{\bbone}{{\mathds{1}}}   

\newcommand{\norm}[1]{\lVert#1\rVert}

\newcommand{\brck}[1]{\left[#1\right]}
\newcommand{\brce}[1]{\left\{#1\right\}}

\newcommand{\indep}{\mathrel{\perp\!\!\!\perp}}

\usepackage[round]{natbib}

\title{\bf Active Measuring in Reinforcement Learning With Delayed Negative Effects}
\author[1]{Daiqi Gao}
\author[2]{Ziping Xu}
\author[1]{Aseel Rawashdeh}
\author[3]{Predrag Klasnja}
\author[1]{Susan A. Murphy}
\affil[1]{Harvard University}
\affil[2]{University of North Carolina at Chapel Hill}
\affil[3]{University of Michigan}
\date{}

\begin{document}

\maketitle





\begin{abstract}
Measuring states in reinforcement learning (RL) can be costly in real-world settings and may negatively influence future outcomes. We introduce the Actively Observable Markov Decision Process (AOMDP), where an agent not only selects control actions but also decides whether to measure the latent state. The measurement action reveals the true latent state but may have a negative delayed effect on the environment. We show that this reduced uncertainty may provably improve sample efficiency and increase the value of the optimal policy despite these costs. We formulate an AOMDP as a periodic partially observable MDP and propose an online RL algorithm based on belief states. To approximate the belief states, we further propose a sequential Monte Carlo method to jointly approximate the posterior of unknown static environment parameters and unobserved latent states. We evaluate the proposed algorithm in a digital health application, where the agent decides when to deliver digital interventions and when to assess users’ health status through surveys.
\end{abstract}

\section{Introduction}

Reinforcement learning (RL) in domains such as games often assumes that the states and rewards are fully observable.
In many real-world applications, however, measuring the states and rewards can be costly and may affect state transitions.
For example, in digital health, an RL algorithm decides when to send  intervention nudges to help users alleviate depression.
To adapt these interventions effectively, the algorithm must measure users’ emotions through ecological momentary assessment (EMA) \citep{targum2021ecological}. Yet frequent assessments impose burden on users and may reduce user engagement and intervention effectiveness in the longer term.
In robotics, a robot exploring an unknown map may need to activate energy-intensive sensors to improve its understanding of the environment, but doing so drains battery power quickly and may limit the exploration range \citep{choudhury2020adaptive}.

In such problems, actions naturally have two components. Control actions (e.g., sending digital interventions or robot moves) affect environment transitions and are optimized to maximize the cumulative rewards as in the standard RL setting.
Measurement actions (e.g., sending surveys or activating sensors) reveal the latent state for a better control action decisions, but may negatively affect future states.

We model this class of problems as an Actively Observable Markov Decision Process (AOMDP), an extension of the Partially Observable Markov Decision Process (POMDP). Without measurement actions, an AOMDP reduces to a standard POMDP where the latent state is costly to observe and emissions are always passively available. When a measurement action is taken, the emission includes the true latent state, though this can introduce delayed negative effects on future states.
The reward in our formulation is a deterministic function of the next latent and observed states, which means it may itself be unobserved. Our framework highlights the tradeoff between the immediate benefits of collapsing state uncertainty and the potential delayed negative impact of measurement.

\paragraph{Main Contributions.}
First, we propose the AOMDP framework to formalize problems where measuring fully resolves state uncertainty but may negatively affect future states. We prove that any tabular AOMDP can be learned with polynomial samples, in contrast to general POMDPs that may require exponential sample complexity. Further, we carefully characterize the trade-off between benefits of state uncertainty reduction and potential negative effects into the future states.


Second, we formulate AOMDP as a periodic POMDP with period length two to address the different state and action spaces when deciding the measurement action and the control action. We propose an online RL algorithm based on the corresponding periodic belief MDPs. The algorithm adapts Randomized Least-Squares Value Iteration (RLSVI) to handle both control and measurement actions, making it lightweight and suitable for settings with limited data.

Third, to obtain the belief state in an unknown environment, we develop a sequential Monte Carlo (SMC) method \citep{del2006sequential} to approximate the joint posterior of \textit{unknown static environment parameters} and \textit{unobserved stochastic latent states}.
A key insight is that the observed state can be viewed an emission of the previous latent state, thus helping update the weights of the particle trajectory.

Finally, we apply the proposed algorithm in a 
digital health application for promoting physical activity, where an RL agent decides when to send intervention nudges and when to query users about their latent commitment to being active.

\section{Related Work}

Active learning in RL strategically selects the most informative actions or states to explore in order to improve learning efficiency and performance.
Active reward learning minimizes the number of reward queries while ensuring a near-optimal policy \citep{daniel2014active,kong2022provably}.
Information-directed reward learning selects a query to provide to the expert at each query time defined by a fixed schedule to maximize the return \citep{lindner2021information}.
Active queries in RL from human feedback selects the conversations or experts to query in order to increase query efficiency \citep{das2024active,ji2024reinforcement,liu2024dual}.

One line of work maximizes the cumulative reward by balancing the reward of the control action and the cost of the measurement action, where the reward and cost are measured in the same unit.
Although the reward in RL can be viewed as a deterministic function of the next state, there is a difference between actively measuring latent states and measuring latent rewards.
When the reward is latent, the probability of measuring will always converge to zero as the estimation of the expected reward becomes more accurate.
However, when the state is latent, the probability of measuring may not converge to zero even if the transition model is known or well learned.
Due to the stochasticity in state transitions, it is not possible to accurately predict the state needed for selecting the action.
Several works \citep{krueger2016active,schulze2018active,tucker2023bandits,parisi2024beyond} focused on latent rewards, while we consider the problem with both latent states and latent rewards, with latent rewards modeled as part of the next latent state.

To actively measure the latent state, \citet{nam2021reinforcement} formally proposed the Action-Contingent Noiselessly Observable MDPs (ACNO-MDPs) framework, which formulated the problem as a special case of a POMDP.
The cost of their measurement action was fixed and observed along with the reward.
However, in AOMDP, the negative effect is delayed and incorporated into future states.
Further, an ACNO-MDP did not allow unobserved rewards or always-passively-observed states and emissions.
\citet{nam2021reinforcement} proposed algorithms for both tabular and continuous settings, but their deep RL algorithm for continuous settings was not feasible for problems with limited data, e.g., in many digital health applications.
Moreover, the estimated transition parameters and latent states were not guaranteed to be drawn from their posterior distributions (see a detailed discussion in Appendix~\ref{sec:additional.related.work}).
\citet{krale2023act,krale2024robust,avalos2024online} proposed lightweight algorithms for solving ACNO-MDPs, but they only considered tabular settings.

In a mixed observability MDP (MOMDP) \citep{ong2010planning}, part of the state is always observed, while the rest is always latent. 
AOMDP can be viewed as a special case of MOMDP, with an extra measurement action that reveals the true latent reward.
\citet{sinha2024periodic} developed a periodic policy for a POMDP, but their underlying environment is stationary.


\section{Problem Setup} \label{sec:problem.setup}

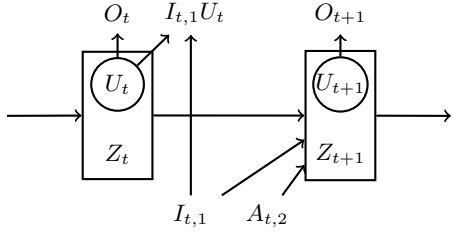
\begin{figure}[t]
    \centering
        \begin{tikzpicture}[->, thick, main/.style={font=\sffamily}]
        \tikzstyle{fixedwidth} = [font=\small, draw=none, text width=0.7cm, align=center]
        \tikzstyle{nodecircle} = [font=\small, draw, circle, minimum size=0.7cm, inner sep=0]
        \matrix [column sep=0.04cm, row sep=0.3cm] {
            & 
            & \node[fixedwidth] (O1) {$O_{t}$};
            & \node[fixedwidth] (IU1) {$I_{t, 1} U_{t}$};
            & & 
            & \node[fixedwidth] (O2) {$O_{t + 1}$}; 
            & \\
            \node[font=\small, draw=none, minimum size=0.7cm, inner sep=0] (U0) {}; 
            & 
            & \node[nodecircle] (U1) {$U_{t}$}; 
            & & & 
            & \node[nodecircle] (U2) {$U_{t + 1}$}; 
            & 
            & \node[font=\small, draw=none, minimum size=0.7cm, inner sep=0] (U3) {}; 
            \\
            \node[fixedwidth, opacity=0] (Z0) {$Z_{t-1}$}; 
            & \node[fixedwidth] (X0) {}; 
            & \node[fixedwidth] (Z11) {$Z_{t}$}; 
            & & 
            & 
            & \node[fixedwidth] (Z21) {$Z_{t + 1}$};
            & \node[fixedwidth] (Z22) {}; 
            & \node[fixedwidth, opacity=0] (Z31) {$Z_{t+1}$}; 
            \\
            & &
            & \node[fixedwidth] (I1) {$I_{t, 1}$}; 
            & 
            & \node[fixedwidth] (A1) {$A_{t, 2}$}; 
            & 
            & \\
        };
        \node[draw=none, rectangle, fit=(U0) (Z0), inner xsep=-0.5pt, inner ysep=2pt] (recZ0) {};
        \node[draw, rectangle, fit=(U1) (Z11), inner xsep=-0.5pt, inner ysep=2pt] (recZ11) {};
        \node[draw, rectangle, fit=(U2) (Z21), inner xsep=-0.5pt, inner ysep=2pt] (recZ21) {};
        \node[draw=none, rectangle, fit=(U3) (Z31), inner xsep=-0.5pt, inner ysep=2pt] (recZ31) {};
        \path[every node/.style={font=\sffamily}]
            (recZ0) edge (recZ11)
            (recZ11) edge (recZ21)
            (recZ21) edge (recZ31)
            (I1) edge (recZ21)
            (A1) edge (recZ21)
            (U1) edge (O1)
            (U2) edge (O2)
            (U1) edge (IU1)
            (I1) edge (IU1)
        ;
        \end{tikzpicture}
    \caption{The diagram showing the environment of an AOMDP. A directed edge connected to a square indicates edges to each node within the square. Edges pointing to the control ($A_t$) and measurement actions ($I_t$) 
    are omitted.}
    \label{fig:dag.general}
\end{figure}

In active measuring, the agent interacts with the environment following the dynamic depicted in Figure \ref{fig:dag.general}. 
At each time step $t$, the agent observes state $Z_t$ and emission $O_t$ (not the latent state $U_t$), and decides the measurement action $I_{t, 1}$. 
Taking $I_{t, 1} = 1$ will reveal the true latent state $U_{t}$, while taking $I_{t, 1} = 0$ means that $U_t$ remains latent.
We view both $O_{t}$ and $I_{t, 1} U_{t}$ as emissions of $U_{t}$.
Then, the agent makes decision about the control action $A_{t, 2}$, and the environment generates the next state $(Z_{t+1}, U_{t+1}) \sim \bbT(\cdot \mid Z_t, U_t, I_{t, 1}, A_{t, 2})$ and the next emission $O_{t + 1} \sim \bbO(\cdot \mid U_{t + 1})$.
The definition of $\bbT$ guarantees that the transition to the next states $Z_{t + 1}, U_{t + 1}$ does not depend on the history prior to time $t$.
The reward after taking $A_{t, 2}$ is $R_t = r(Z_{t+1}, U_{t+1})$, where $r$ is known deterministic function. Since the reward depends on the latent state $U_{t}$, the reward may also be latent. 
Note that there is no instantaneous reward at time $(t, 1)$ for $I_{t, 1}$, that is between $I_{t, 1}$ and $A_{t, 2}$). Thus the effect of $I_{t, 1}$ on rewards is only via future states, $Z_{t + 1}, U_{t+1}$. 
While the control action $A_{t, 2}$ only affects the transition $\bbT$, the measurement action $I_{t, 1}$ affects both the transition $\bbT$ and the emission $I_{t, 1} U_{t}$.
The observed history before choosing $I_{t, 1}$ is $H_{t, 1} = \{Z_{1}, O_{1}, I_{1, 1}, I_{1, 1} U_{1}, A_{1, 2}, \dots, Z_{t}, O_{t}\}$.
The observed history before choosing $A_{t, 2}$ is $H_{t, 2} = H_{t, 1} \cup \{I_{t, 1}, I_{t, 1} U_{t}\}$.

Definition~\ref{def:aomdp} formally defines active measuring with delayed effects on the environment as a POMDP with mixed observable states and special emission structures.  
For a set $\cS$, let $\Delta(\cS)$ be the set of probability measures on the measurable space $(\cS, \cB(\cS))$, where $\cB(\cS)$ is the Borel $\sigma$-algebra on $\cS$.

\begin{dfn}[AOMDP] \label{def:aomdp}
    An \textit{Actively Observable MDP} is a tuple $(\cZ, \cU, \cO, \cA, \cI, \bbT, \bbO, r, \gamma)$, where
    $\cA$ is the control action space, 
    $\cI = \{0, 1\}$ is the measurement action space,
    $\cZ \subseteq \bbR^{d_{Z}}$ is the space of observed states,
    $\cU \subseteq \bbR^{d_U}$ is the space of latent states,
    and $\cO \subseteq \bbR^{d_O}$ is the space of emissions of $\cU$.
    The emission function is $\bbO: \cU \mapsto \Delta(\cO)$, with $\bbO (o_{t} \mid u_{t})$ being the probability density function (p.d.f.) of $O_{t}$ given $U_{t} = u_{t}$ (overloading notation).
    The transition function for $Z_{t}$ and $U_{t}$ is $\bbT: \cZ \times \cU \times \cI \times \cA \mapsto \Delta(\cZ \times \cU)$, with $\bbT (z_{t+1}, u_{t+1} \mid z_{t}, u_{t}, i_{t, 1}, a_{t, 2})$ being the p.d.f. of $Z_{t+1}, U_{t+1}$ given $Z_{t} = z_{t}, U_{t} = u_{t}, I_{t, 1} = i_{t, 1}, A_{t, 2} = a_{t, 2}$.
    The reward function $r: \cZ \times \cU \mapsto \bbR$ is a known deterministic function of the next observed and latent states. 
    The discount factor $\gamma \in [0, 1)$ is a constant.
    We assume that $Z_{t}$ and $O_{t}$ are observed before choosing $I_{t, 1}$, and that $I_{t, 1} U_{t}$ is observed before choosing $A_{t, 2}$.
\end{dfn}


The goal is to find a policy $\bpi$ that selects $I_{t, 1}$ and $A_{t, 2}$ to maximize the expected discounted sum of rewards $\bpi^* = \argmax_{\bpi} \bbE^{\bpi} \{ \sum_{t=1}^{\infty} \gamma^{t - 1} R_{t} \}$.
The positive effect of taking the measurement action $I_{t, 1} = 1$ is twofold: (1) learning: it helps learn the transition and emission functions, either directly via SMC or indirectly in a model-free RL algorithm; and (2) optimization: it provides an accurate state for choosing $A_{t, 2}$. This second benefit exists even when the environment is known.
A general problem with no information of the reward will be impossible to solve, since an agent cannot learn from any feedback.
However, the emissions $O_{t}$ and $I_{t, 1} U_{t}$ and the next observed state $Z_{t + 1}$ will help infer the latent state $U_{t}$ and latent reward $R_{t}$.

\subsection{AOMDP and Periodic POMDP}
\label{belief.state}

One major difference between AOMDPs and stationary POMDPs is that the state, action, and emission spaces are time-inhomogeneous at the step $(t,1)$ and step $(t, 2)$, while being periodic on a higher level index $t$. 
This structure is a special case of a periodic POMDP (see definition in Appendix~\ref{sec:periodic.belief.mdp}).
A periodic POMDP allows non-stationarity within a period but assumes the period structure is homogeneous over time.  

\begin{lem} \label{lem:aomdp.periodic.pomdp}
    An AOMDP is a special case of a periodic POMDP with period length $K = 2$. Further, at time $k = 1$, the emission of the state $S^I_{t, 1} = [Z_{t}, U_{t}]$ is $O^I_{t, 1} = [Z_{t}, O_{t}]$, and the reward is zero. 
    At time $k = 2$, the emission of the state $S^A_{t, 2} = [Z_{t}, U_{t}, I_{t, 1}]$ is $O^A_{t, 2} = [Z_{t}, O_{t}, I_{t, 1} U_{t}, I_{t, 1}]$, and the reward is $R_t$. The discount factors are $\gamma_1 = \gamma_2 = \sqrt{\gamma}$.
\end{lem}


A periodic POMDP can be viewed as a stationary POMDP by augmenting the state with the time index $k$, similar as how a periodic MDP is viewed as a stationary MDP \citep{riis1965discounted,sinha2024periodic}.
As discussed in  \cite{kaelbling1998planning}, a POMDP can be solved using a belief MDP, whose optimal policy is a Markov stationary policy based on the belief state.
Here, a belief state is a probability measure that represents the posterior distribution of the latent state given the observed history, and can be viewed as a sufficient statistic of the history.
Incorporating time dependency, the optimal policy of a $K$-periodic POMDP is a sequence of $K$ Markov policies based on the belief state, and can be solved using a periodic belief MDP (see definition in Appendix~\ref{sec:periodic.belief.mdp}).

Concretely, for the measurement action in an AOMDP, the belief state of a latent state $S^I_{t, 1} = [Z_{t}, U_{t}] \in \cS^I$ given the history $H_{t, 1}$ is $b^{S^I}_{t, 1} \in \cB^I$, where $\cB^I = \Delta (\cS^I)$ is the set of belief states over $\cS^I$.  
In other words, $b^{S^I}_{t, 1} (s^I) = p (s^I \mid H_{t, 1})$  is the p.d.f. of the posterior distribution of $S^I_{t, 1}$ given $H_{t, 1}$. 
When the observed state is $Z_{t} = z_{t}$, we have $b^{S^I}_{t, 1} = \delta_{z_{t}} \otimes b^U_{t, 1}$, where $\delta$ is the Dirac measure, $b^U_{t, 1}$ is the belief state of $U_{t}$ at time $(t, 1)$, and $\mu_1 \otimes \mu_2$ denotes the product measure of $\mu_1$ and $\mu_2$. 
Similarly, for the control action, the belief state of a latent state $S^A_{t, 2} = [Z_{t}, U_{t}, I_{t, 1}] \in \cS^A$ given the history $H_{t, 2}$ is $b^{S^A}_{t, 2} \in \cB^A$, where $\cB^A = \Delta (\cS^A)$ and $b^{S^A}_{t, 2} (s^A) = p (s^A \mid H_{t, 2})$.
When the observed state is $Z_{t} = z_{t}$ and the measurement action is $I_{t, 1} = i_{t, 1}$, we have $b^{S^A}_{t, 2} = \delta_{z_{t}} \otimes b^U_{t, 2} \otimes \delta_{i_{t, 1}}$, where $b^U_{t, 2}$ is the updated belief state of $U_{t}$ after observing $I_{t, 1} U_{t}$.  
We will discuss how to estimate the belief state $b^U_{t, 1:2}$ of $U_{t}$ when $\bbT$ is unknown in Section~\ref{sec:belief.propagation}.

Lemma~\ref{lem:aomdp.periodic.pomdp} implies that the AOMDP can be solved as a periodic belief MDP with $K = 2$.  
Thus, the optimal policy of an AOMDP is Markov stationary policy $\bpi := \{\pi^I, \pi^A\}$ with $\pi^I: \cB^I \mapsto \cI$ and $\pi^A: \cB^A \mapsto \cA$.
The reward after taking $A_{t, 2}$ based on the belief state is $r (Z_{t + 1}, b^{U}_{t + 1}) = r (b^{S^I}_{t + 1, 1}) = \int r (s) b^{S^I}_{t + 1, 1} (s) ds$.
Then, the Q-functions of $A$ and $I$ are defined as
\begin{align*}
    \cQ^{I \bpi} (b^{S^I}_{t, 1}, i_{t, 1}) &:= 
    \bbE^{\bpi} \brce{
    \sum_{l = t}^{\infty} 
    \gamma^{l - t} r (b^{S^I}_{l + 1, 1})
    \,\middle|\, b^{S^I}_{t, 1}, i_{t, 1} }, \\
    \cQ^{A \bpi} (b^{S^A}_{t, 2}, a_{t, 2}) &:= 
    \bbE^{\bpi} \brce{
    \sum_{l = t}^{\infty} 
    \gamma^{l - t - \frac{1}{2}} r (b^{S^I}_{l + 1, 1})
    \,\middle|\, b^{S^A}_{t, 2}, a_{t, 2} }.
\end{align*}
The Bellman optimality equations for the AOMDP is
\begin{align}
    \cQ^{I*} (b^{S^I}_{t, 1}, i_{t, 1}) &= 
    \bbE \Big\{
    \sqrt{\gamma} \max_{a \in \cA}
    \cQ^{A*} (b^{S^A}_{t, 2}, a) 
    \,\Big|\, b^{S^I}_{t, 1}, i_{t, 1} \Big\},
    \label{equ:bellman.optimality.I} \\
    \cQ^{A*} (b^{S^A}_{t, 2}, a_{t, 2}) &= 
    \bbE \Big\{r (b^{S^I}_{t + 1, 1}) + 
    \sqrt{\gamma} \max_{i \in \cI} 
    \cQ^{I*} (b^{S^I}_{t + 1, 1}, i) 
    \,\Big|\, b^{S^A}_{t, 2}, a_{t, 2} \Big\},
    \label{equ:bellman.optimality.A}
\end{align}
which is extended from results of periodic belief MDPs.


\section{Methodology} \label{sec:methodology}

To learn the optimal Q-function $\cQ^*_k$ online, we adapt the RLSVI algorithm \citep{osband2016generalization} to the periodic belief MDP based on the Bellman optimality equations~\eqref{equ:bellman.optimality.I} and~\eqref{equ:bellman.optimality.A}.  
RLSVI is a model-free algorithm that selects the greedy action with respect to a sample of the policy parameter drawn from its posterior distribution (see details in Appendix~\ref{sec:rlsvi}).  
In order to obtain the belief states $b^{S^I}_{t, 1}$ and $b^{S^A}_{t, 2}$, we assume parametric transition and emission models in Section~\ref{sec:belief.propagation}.  
The use of a model-free RLSVI algorithm provides  robustness against misspecification in these parametric models. 

We approximate each optimal Q-function by a linear function of the basis function $\phi$, i.e.
\begin{equation} \label{equ:linear.definition.Q}
    \begin{split}
        \cQ^{I*} (b^{S^I}_{t, 1}, I_{t, 1}) &= \phi^I (b^{S^I}_{t, 1}, I_{t, 1})^{\top} \bbeta^I, \\
        \cQ^{A*} (b^{S^A}_{t, 2}, A_{t, 2}) &= \phi^A (b^{S^A}_{t, 2}, A_{t, 2})^{\top} \bbeta^A,
    \end{split}
\end{equation}
where $\phi^I$ and $\phi^A$ are the basis functions, and $\bbeta^I$ and $\bbeta^A$ are the parameters to be learned.  
For example, the basis can be of linear, polynomial, or Gaussian functions of the state and action.  
For this choice of basis functions, RLSVI fits a Bayesian linear regression (BLR) model to the target, which is the estimated optimal Q-function.

\subsection{Approximating the Belief State} \label{sec:belief.propagation}

\begin{algorithm}[t]
    \caption{Estimating Belief State $b^U_{t, 1}$}
    \label{alg:belief.state.1}
    \textbf{Input}: history $h_{t, 1}$, $J$ particles $\{ \widehat{u}_{1:t-1, 1:2}^{(j)} \}_{j=1}^J$ with weights $\{ w_{t-1, 2}^{(j)} \}_{j=1}^J$, and the prior of $\btheta$.
    
    \begin{algorithmic}[1] 
    \FOR{$j \in \{1:J\}$}
        \STATE Draw $\widehat{\btheta}_t^{(j)} \sim p (\btheta \mid \widehat{u}_{1:t-1, 2}^{(j)}, h_{t-1,2})$, $\widetilde{u}_{t, 1}^{(j)} \sim p ( u_{t} \mid z_{t}, \widehat{\btheta}_t^{(j)}, \widehat{u}_{t-1, 2}^{(j)}, z_{t-1}, i_{t - 1, 1}, a_{t - 1, 2} )$. 
        \STATE Particle weight:
        $ \widetilde{w}_{t, 1}^{(j)} \propto w_{t-1, 2}^{(j)} p ( z_{t} \mid \widehat{\btheta}_t^{(j)}, \allowbreak \widetilde{u}_{t-1, 2}^{(j)}, \allowbreak z_{t-1}, \allowbreak i_{t - 1, 1}, a_{t - 1, 2} ) p ( o_{t} \mid \widehat{\btheta}_t^{(j)}, \widetilde{u}_{t, 1}^{(j)} )$.  
        \STATE Calculate the effective sample size $\text{ESS} := [\sum_{i=1}^N (\widetilde{w}_{t, 1}^{(j)})^2]^{-1}$.  
        If $\text{ESS} < 0.5 J$, resample from $\{ \widehat{u}_{1:t-1, 1:2}^{(j)}, \widetilde{u}_{t, 1}^{(j)} \}_{j=1}^J$ with weights $\{ \widetilde{w}_{t, 1}^{(j)} \}_{j=1}^J$ to obtain $J$ new particles $\{ \widehat{u}_{1:t-1, 1:2}^{(j)}, \widehat{u}_{t, 1}^{(j)} \}_{j=1}^J$ with weights $w_{t, 1}^{(j)} = 1/J$ for all $j$.  
        Otherwise, set $\widehat{u}_{t, 1}^{(j)} = \widetilde{u}_{t, 1}^{(j)}$ and $w_{t, 1}^{(j)} = \widetilde{w}_{t, 1}^{(j)}$.  
    \ENDFOR
    \STATE The estimated belief state is $\widehat{b}^U_{t, 1} (u) = \sum_{j=1}^J w_{t, 1}^{(j)} \delta (u - \widehat{u}_{t, 1}^{(j)})$.
    \end{algorithmic}
\end{algorithm}

In this section, we generalize the Particle Belief MDP (PB-MDP) approximation~\citep{lim2023optimality} to the AOMDP.
The PB-MDP 
approximates the belief states by using SMC to maintain a set of $J$ particles.  

Generalizing the PB-MDP approximation first requires addressing the challenge that neither the transition function $\bbT$ nor the emission function $\bbO$ is known.  
To remedy this here we use parametric models; denote $\btheta$ as the joint parameters of the transition and emission functions $\bbT, \bbO$.  
The second challenge is that  augmenting the latent state with the static parameter of the environment, $\btheta$, in standard particle filtering fails, since the parameter space is only explored in the initial step and its posterior distribution degenerates as time increases \citep{kantas2015particle}.
We  use ideas from particle learning
\citep{storvik2002particle,carvalho2010particle}, which samples a new $\btheta$ at every step.  Our  solution will enable efficient online SMC with static parameter estimation since, under some working models (detailed in Section~\ref{sec:posterior.parameters}), the posterior of $\btheta$ given the history and fixed values of the latent states is  closed-form.  
Finally, the belief state of $U_t$ is updated at both $(t, 1)$ and $(t, 2)$.  

In the AOMDP, we only need to derive the belief states $b^{U}_{t, 1:2}$ for $U_t$ to construct $b^{S^I}_{t, 1}$ and $b^{S^A}_{t, 2}$ (discussed in Section~\ref{belief.state}).
In SMC, each particle represents a possible trajectory of latent states $U_{1:t}$ up to the current time $t$.  
The marginal posterior of the current latent state $U_t$  given the history $H_{t, 1}$ or $H_{t, 2}$ is approximated by the empirical distribution of the last state in each particle $\{U_t^{(j)}\}_{j=1}^J$.  
The particles are updated at each time step based on the newly observed data.

\begin{algorithm}[t]
    \caption{Estimating Belief State $b^U_{t, 2}$}
    \label{alg:belief.state.2}
    \textbf{Input}: history $h_{t, 2}$, $J$ particles $\{ \widehat{u}_{1:t-1, 2}^{(j)}, \widehat{u}_{t, 1}^{(j)} \}_{j=1}^J$ with weights $\{ w_{t-1}^{(j)} \}_{j=1}^J$.
    
    \begin{algorithmic}[1] 
    \IF{$i_{t, 1} = 1$}
        \STATE When $I_{t, 1} U_{t} = u_t$, set $\widehat{u}_{t, 2}^{(j)} = u_{t}$ for all $j \in \{1:J\}$.  
        Update the particle weight $w_{t, 2}^{(j)} = w_{t-1, 2}^{(j)} p ( z_{t}, u_t \mid \widehat{\btheta}_t^{(j)}, u_{t-1, 2}^{(j)}, z_{t - 1}, i_{t - 1, 1}, a_{t - 1, 2} )$.  
    \ELSE
        \STATE Set $\widehat{u}_{t, 2}^{(j)} = \widehat{u}_{t, 1}^{(j)}$ for each $j \in \{1:J\}$.  
        Update the particle weight $w_{t, 2}^{(j)} = w_{t, 1}^{(j)}$.  
    \ENDIF
    \STATE The estimated belief state is $\widehat{b}^U_{t, 2} (u) = \sum_{j=1}^J w_{t, 2}^{(j)} \delta (u - \widehat{u}_{t, 2}^{(j)})$.
    \end{algorithmic}
\end{algorithm}


Leveraging the idea of particle learning \citep{storvik2002particle,carvalho2010particle}, at each time $t$ we first draw the parameter $\widehat{\btheta}_t^{(j)}$ from its posterior given the observed history $h_{t, k}$ and the value of one particle $\widehat{u}_{1:t-1, 2}^{(j)}$ up to time $t - 1$, before drawing the new state $U_t^{(j)}$ given $\widehat{\btheta}_t^{(j)}$, $h_{t, k}$, and $\widehat{u}_{1:t-1, 2}^{(j)}$.  
See Algorithms~\ref{alg:belief.state.1} and~\ref{alg:belief.state.2}.  Note that Algorithm~\ref{alg:belief.state.1} uses an explicit  formula for  the posterior of $\btheta$ given the observed history and fixed values of the latent states.   
Next, notice that $p ( z_{t} \mid \btheta, u_{t-1}^{(j)}, z_{t-1}, i_{t - 1, 1}, a_{t - 1, 2} )$ is used to approximate the belief state $\widehat{b}^U_{t, 1}$ in Algorithm~\ref{alg:belief.state.1}, even though it does not involve $U_{t}$.  
This is because each particle is a draw from the posterior of the latent state trajectory. $Z_t$ acts as an emission of the latent state trajectory. 
Indeed a small likelihood of $Z_{t}$ indicates that the previous latent state value $U_{t-1}^{(j)}$ is less likely to be the true value, and this trajectory should therefore be down-weighted.  
Further, when $i_{t, 1} = 1$, to avoid the case where all weights $w_{t, 2}^{(j)} = 0$ if the true value $u_t \notin \{\widehat{u}_{t, 1}^{(j)} \}_{j=1}^J$, we resample the particles and update the weights from $w_{t-1, 2}^{(j)}$.
Lastly, Algorithm~\ref{alg:belief.state.1} resamples the whole trajectory when the ESS is small for better numerical stability \citep{liu1998sequential,del2006sequential}.
The derivation of the sampling scheme in Algorithms~\ref{alg:belief.state.1} and~\ref{alg:belief.state.2} is provided in Appendix~\ref{sec:proof.belief.state}.  
Denote the approximate posterior distributions from Algorithms~\ref{alg:belief.state.1} and~\ref{alg:belief.state.2} by  $\widehat{b}^U_{t, 1}$ and $\widehat{b}^U_{t, 2}$ respectively.  The  belief states of $S^I_{t, 1}$ and $S^A_{t, 2}$ are approximated by $\widehat{b}^{S^I}_{t, 1} = \delta_{z_{t}} \otimes \widehat{b}^U_{t, 1}$ and $\widehat{b}^{S^A}_{t, 2} = \delta_{z_{t}} \otimes \widehat{b}^U_{t, 2} \otimes \delta_{i_{t, 1}}$.

\subsection{Constructing the Target} \label{sec:target}

Motivated by the fact that the control and measurement actions have different effects on the environment, we construct the targets for them differently.  
When $b^{S^I}_{t, 1} = \delta_{z_{t}} \otimes b^U_{t, 1}$, the target for the measurement action based on Equation \eqref{equ:bellman.optimality.I} can be rewritten as
\begin{align}
    \cQ^{I*} (b^{S^I}_{t, 1}, 1)
    &= \sqrt{\gamma} \int \max_{a \in \cA} Q^{A*} ( \delta_{z_t} \otimes \delta_{u_t} \otimes \delta_{1}, a ) b^{U}_{t, 1} (u_t) d u_t, \label{equ:Q.I1.conclusion} \\
    \cQ^{I*} (b^{S^I}_{t, 1}, 0)
    &= \sqrt{\gamma} \max_{a \in \cA} Q^{A*} ( \delta_{z_t} \otimes b^{U}_{t, 1} \otimes \delta_{0}, a ) \label{equ:Q.I0.conclusion}.
\end{align}
Note that \eqref{equ:Q.I1.conclusion} is an integration over the distribution of emission $I_{t, 1} U_t$.
This is derived based on the known transition function from $b^{U}_{t, 1}$ to $b^{U}_{t, 2}$ (see the proof in Appendix~\ref{sec:target.I}).  
When $b^{S^I}_{t, 1} = \delta_{z_{t}} \otimes b^U_{t, 1}$, $I_{t, 1} = 1$, the emission $I_{t, 1} U_{t}$ has a p.d.f. $b^U_{t, 1}$. Given $I_{t, 1} U_{t} = u_t$, we have $b^{S^A}_{t, 2} (s^A) = \delta_{z_t} \otimes \delta_{u_t} \otimes \delta_{1}$.
When $b^{S^I}_{t, 1} = \delta_{z_{t}} \otimes b^U_{t, 1}$, $I_{t, 1} = 0$, we have $P (I_{t, 1} U_{t} = 0) = 1$ and $b^{S^A}_{t, 2} (s^A) = \delta_{z_t} \otimes b^{U}_{t, 1} \otimes \delta_{0}$.
Further, since the instantaneous reward for the measurement action is zero, 
the target for the control action is constructed based on a \textit{2-step TD prediction} \citep[Chapter 7]{sutton2018reinforcement}.  
This allows the target of the control policy to be updated based on itself rather than the measure policy, which we find improves numerical stability. 
In addition, we use double Q-learning  to alleviate the maximization bias in the targets.  This is essential for the active measuring target.  
Notice that if there were no delayed  effect of measurement actions, the second benefit of measuring (obtaining an accurate state) comes exactly from the difference between \eqref{equ:Q.I1.conclusion} and \eqref{equ:Q.I0.conclusion} (see details in Proposition~\ref{prop:measure.advantage}), which could be significantly overestimated due to maximization bias.  

Based on the above discussion, let $\bX^I_{l} = \phi^I (\widehat{b}^{S^I}_{l, 1}, I_{l, 1})$ and $\bX^A_{l} = \phi^A (\widehat{b}^{S^A}_{l, 2}, A_{l, 2})$ 
be the covariates in the BLR for $l \in \{1:t\}$. 
When $I_{l, 1} = 1$, define the target for the measurement action as 
\begin{equation*}
    \begin{split}
        Y^I_{l} = \sqrt{\gamma} \sum_{j=1}^J w_{t, 1}^{(j)} [ \phi^A ( \delta_{z_t} \otimes \delta_{\widehat{u}_{t, 1}^{(j)}} \otimes \delta_{1}, a^{(j)} )^{\top} \widetilde{\bbeta}^A_{t - 1} ], \hspace{1em} 
        \text{where }
        a^{(j)} = \argmax\limits_{a \in \cA} [ \phi^A ( \delta_{z_t} \otimes \delta_{\widehat{u}_{t, 1}^{(j)}} \otimes \delta_{1}, a )^{\top} \widetilde{\bbeta}^A_{t^-} ],
    \end{split}
\end{equation*}
since $\widehat{b}^U_{t, 1} (u) = \sum_{j=1}^J w_{t, 1}^{(j)} \delta (u - \widehat{u}_{t, 1}^{(j)})$,
and when $I_{l, 1} = 0$, define  
\begin{equation*}
    \begin{split}
        Y^I_{l} = \sqrt{\gamma} \phi^A (\delta_{z_t} \otimes \widehat{b}^{U}_{t, 1} \otimes \delta_{0}, a')^{\top} \widetilde{\bbeta}^A_{t - 1}, \hspace{1em} 
        \text{where }
        a' = \argmax\limits_{a \in \cA} [ \phi^A (\delta_{z_t} \otimes \widehat{b}^{U}_{t, 1} \otimes \delta_{0}, a)^{\top} \widetilde{\bbeta}^A_{t - } ].
    \end{split}
\end{equation*}
Here, $\widetilde{\bbeta}^A_{t - 1}$ is the estimated parameter at time $t - 1$, and $\widetilde{\bbeta}^A_{t^-}$ is 
copied from $\widetilde{\bbeta}^A_{t}$ every $C$ steps. 
While the standard RLSVI \citep{osband2016generalization} approximates the target with a single observation of the next state to increase computational efficiency when the transition is unknown, we can directly evaluate the expectation in $Y^I_{l}$ to increase numerical stability due to the known transition from $b^{U}_{t, 1}$ to $b^{U}_{t, 2}$.
For the control action $A$, define the target as
\begin{equation*}
    \begin{split}
        Y^A_{l} = r(Z_{l + 1}, \widehat{b}^{U}_{l + 1, 1}) + \gamma \phi^A (\widehat{b}^{S^A}_{l + 1, 2}, a')^{\top} \widetilde{\bbeta}^A_{t - 1}, \hspace{1em} 
        \text{where }
        a' = \argmax_{i \in \cI} \phi^A (\widehat{b}^{S^A}_{l + 1, 2}, i')^{\top} \widetilde{\bbeta}^A_{t^-}.
    \end{split}
\end{equation*}
Based on the definition, the reward can be estimated as $\widehat{R} = r(Z_{t + 1}, \widehat{b}^{U}_{t + 1, 1}) = \sum_{j=1}^J w_{t + 1, 1}^{(j)} r(Z_{t + 1}, \widehat{u}_{t + 1, 1}^{(j)})$.
The construction of $\phi^I$ and $\phi^A$ is problem-dependent. Appendix~\ref{sec:heartsteps.aomdp} describes how $\phi^I$ and $\phi^A$ are constructed for the application in Section~\ref{sec:application}.
For example, when the belief state is approximately normal, we can use the weighted mean and standard deviation of the particles $\{ \widehat{u}_{t, k}^{(j)} \}_{j=1}^J$, $k = 1$ or 2, along with the observed state $Z_t$ to construct the basis functions.

To update the parameter $\bbeta^I$, we fit a BLR on $\Yb^I := [Y^I_{1:t}]^{\top}$ using $\Xb^I := [\bX^I_{1:t}]^{\top}$.  
Similarly, for $\bbeta^A$, we fit a BLR on $\Yb^A = [Y^A_{1:t}]^{\top}$ using $\Xb^A = [\bX^A_{1:t}]^{\top}$.  
The posterior of $\bbeta^I_{t}$ is $N (\bmu^I_{t}, \bSigma^I_{t})$,  
and the posterior of $\bbeta^A_{t}$ is $N (\bmu^A_{t}, \bSigma^A_{t})$,  
where 
\begin{equation} \label{equ:blr.posterior}
    \begin{split}
        \bSigma^I_t &= [(\Xb^I_t)^{\top} \Xb^I_t / (\sigma^I)^2 + \lambda^I \Ib]^{-1}, \\
        \bmu^I_t &= \bSigma^I_t [(\Xb^I_t)^{\top} \Yb^I_t / (\sigma^I)^2], \\
        \bSigma^A_t &= [(\Xb^A_t)^{\top} \Xb^A_t / (\sigma^A)^2 + \lambda^A \Ib]^{-1}, \\
        \bmu^A_t &= \bSigma^A_t [(\Xb^A_t)^{\top} \Yb^A_t / (\sigma^A)^2],
    \end{split}
\end{equation}
and $\Ib$ is the identity matrix.  
An estimate of $\bbeta^I_{t}$ is then obtained by drawing $\widetilde{\bbeta}^I_{t} \sim N (\bmu^I_{t}, \bSigma^I_{t})$ from the posterior distribution.  
Similarly, $\widetilde{\bbeta}^A_{t} \sim N (\bmu^A_{t}, \bSigma^A_{t})$ is drawn.  
Finally, $I_{t, 1}$ and $A_{t, 2}$ are selected by maximizing the estimated Q-functions based on $\widetilde{\bbeta}^I_{t}$ and $\widetilde{\bbeta}^A_{t}$.
See Algorithm~\ref{alg:active.measure} in Appendix~\ref{sec:algorithm.details}.

\begin{figure*}[t]
    \centering
    \begin{tikzpicture}[->, thick, main/.style={font=\sffamily}]
    \tikzstyle{fixedwidth} = [font=\normalsize, draw=none, text width=0.7cm, align=center]
    \tikzstyle{nodecircle} = [font=\normalsize, draw, circle, minimum size=0.8cm, inner sep=0]
    \matrix [column sep=0.4cm, row sep=0.3cm] {
        & \node[fixedwidth] (O0) {$O_{t - 2}$};
        & \node[fixedwidth] (IR0) {$I_{t - 1} R_{t - 2}$};
        & & & 
        & \node[fixedwidth] (O1) {$O_{t - 1}$};
        & \node[fixedwidth] (IR1) {$I_{t} R_{t - 1}$};
        & & & 
        & \node[fixedwidth] (O2) {$O_{t}$}; 
        & \node[fixedwidth] (IR2) {$I_{t + 1} R_{t}$};
        \\
        \node[draw=none, text width=0cm] (R00) {};
        & \node[nodecircle] (R0) {$R_{t - 2}$};
        & & & & 
        & \node[nodecircle] (R1) {$R_{t - 1}$}; 
        & & & & 
        & \node[nodecircle] (R2) {$R_{t}$}; 
        & 
        & \node[fixedwidth] (R3) {}; \\
        & & 
        & 
        & \node[fixedwidth] (C1) {\hspace{-1em}$C_{t - 1, 2}$}; 
        & \node[fixedwidth] (M1) {$M_{t - 1, 2}$};
        & & 
        & 
        & \node[fixedwidth] (C2) {$C_{t, 2}$}; 
        & \node[fixedwidth] (M2) {$M_{t, 2}$};
        & & 
        & \\
        & \node[draw=none, text width=0.1cm] (A0) {}; 
        & \node[fixedwidth] (I0) {$I_{t - 1, 1}$};
        & 
        & \node[fixedwidth] (A1) {$A_{t - 1, 2}$};
        & & 
        & \node[fixedwidth] (I1) {$I_{t, 1}$};
        & 
        & \node[fixedwidth] (A2) {$A_{t, 2}$};
        & & 
        & \node[fixedwidth] (I2) {$I_{t + 1, 1}$};
        & \\
        \node[draw=none, text width=0cm] (E00) {};
        & \node[fixedwidth] (E0) {$E_{t - 2}$};
        & & & & 
        & \node[fixedwidth] (E1) {$E_{t - 1}$}; 
        & & & & 
        & \node[fixedwidth] (E2) {$E_{t}$}; 
        & 
        & \node[fixedwidth] (E3) {}; 
        \\
    };
    \path[every node/.style={font=\sffamily}]
        (R0) edge [color7] (O0)
        (R1) edge [color7] (O1)
        (R2) edge [color7] (O2)
        (R00) edge [color1] (R0)
        (R0) edge [color1] (R1)
        (R1) edge [color1] (R2)
        (R2) edge [color1] (R3)
        (E00) edge [color6] (E0)
        (E0) edge [color6] (E1)
        (E1) edge [color6] (E2)
        (E2) edge [color6] (E3)
        (E0) edge [color6] (R0)
        (E1) edge [color6] (R1)
        (E2) edge [color6] (R2)
        (R0) edge [color2] (M1)
        (M1) edge [color3] (R1)
        (A1) edge [color5] (E1)
        (E0) edge [color4] (M1)
        (E1) edge [color4] (M2)
        (A1) edge (M1)
        (C1) edge (M1)
        (A2) edge (M2)
        (C2) edge (M2)
        (R1) edge [color2] (M2)
        (M2) edge [color3] (R2)
        (A2) edge [color5] (E2)
        (R0) edge [color8] (IR0)
        (I0) edge [color8] (IR0)
        (R1) edge [color8] (IR1)
        (I1) edge [color8] (IR1)
        (R2) edge [color8] (IR2)
        (I2) edge [color8] (IR2)
        (I0) edge [color8] (E1)
        (I1) edge [color8] (E2)
    ;
    \end{tikzpicture}
    \caption{Causal DAG of HeartSteps. Arrows pointing to the actions are omitted.}
    \label{fig:dag}
\end{figure*}
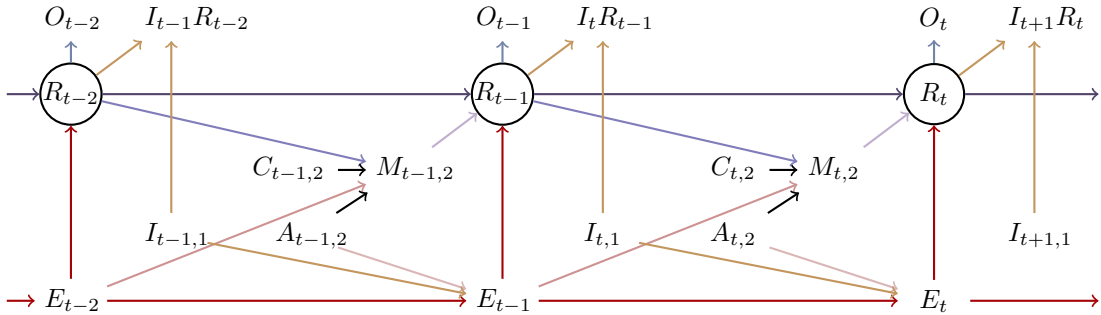

\section{Benefits of Measuring} \label{sec:theoretical.results}

In this section, we discuss the benefits of measurement actions from two aspects—sample complexity benefits and policy improvement benefits—even though they may have negative delayed effects on future cumulative rewards.  
First, measuring may reduce the sample complexity of learning, leading to a more identifiable environment, as it reveals the latent state and thus removes a major source of uncertainty.  
Second, measuring can directly increase the value of the optimal policy by providing state information that enables better decisions in subsequent steps.

\subsection{Sample Complexity Improvement}

We first consider the impact of measuring on the number of samples required to learn an $\epsilon$-optimal policy in an unknown environment.  
In general, learning in POMDPs can be fundamentally challenging: without further assumptions, the sample complexity may grow exponentially with the horizon $H$.  
In finite-horizon tabular POMDPs with always observed rewards, \citet{liu2022partially} introduced the notion of an $m$-step $\alpha$-weakly revealing POMDP (Definition \ref{dfn:alpha_revealing}), which requires that the latent states can be distinguished through $m$-step observations and actions, and thus the system becomes strongly identifiable.  
They showed that under the $m$-step $\alpha$-weakly revealing condition, there exists an algorithm that learns an $\epsilon$-optimal policy with $\operatorname{poly}(S, A^m, H)$ samples, where $S$ is the number of latent states, $A$ is the action space size, and $H$ is the horizon length.  

\begin{dfn}[$m$-step $\alpha$-weakly revealing condition] \label{dfn:alpha_revealing}
There exist $m \in \mathbb{N}$ and $\alpha>0$ such that $\sigma_{S}(M) \geq \alpha$, where for all $(\mathbf{a}, \mathbf{o}) \in \mathcal{A}^{m-1} \times \mathcal{O}^m$ and $s \in \mathcal{S}$,
    \begin{align*}
        \left[M_t \right]_{(\mathbf{a}, \mathbf{o}), s} := \mathbb{P}\left(o_{t: t+m-1}=\mathbf{o} \mid s_t=s, a_{t: t+m-2}=\mathbf{a}\right).
    \end{align*}
    Here, $\sigma_{S}(M_t)$ denotes the $S$-th singular value of the $m$-step emission matrix $M_t$.
\end{dfn}

The following proposition shows that the presence of measurements guarantees a strong form of this condition in the special case of a tabular POMDP.


\begin{prop} \label{prop:AOMDP.alpha.revealing}
    Any finite-horizon tabular AOMDP with the reward $r(Z_{t+1})$ depending only on the observed state satisfies the 2-step 1-weakly revealing condition.
\end{prop}

This result implies that any AOMDP admits polynomial sample complexity, in contrast to general POMDPs without measurement actions.

\subsection{Optimal Policy Value Improvement}

Beyond the learning efficiency benefits, we consider the case where the environment dynamics are fully known.  
Even in this setting, measurement actions can strictly increase the value of the optimal policy by reducing uncertainty about the latent state $U_t$, thereby enabling more informed control decisions.  
Specifically, a measure collapses the belief distribution $b^U_{{t,1}}$ (the posterior over $U_t$ before measuring) to a Dirac measure at the true latent state, while not measuring forces the agent to act under uncertainty.  

To clearly characterize the improvement, we define $\cV_{z, i}^{A*}(b^{U}) := \max_a \cQ^{A*}(\delta_{z} \otimes b^{U} \otimes \delta_i, a)$ for $i \in \{0, 1\}$ as the optimal value function of the control action under the measurement action being 0 and 1, respectively.  

\begin{prop}
    \label{prop:measure.advantage}
    At belief state $b^{S^I} := \delta_z \otimes b^U$, the advantage of measuring is 
    \begin{equation*}
         \cQ^{I*}(b^{S^I}, 1) - \cQ^{I*}(b^{S^I}, 0) 
        = \underset{\textcircled{1}\text{Delayed effect}}{\mathbb{E}_{b^U} [\cV_{z, 1}^{A*}(\delta_{u}) - \cV_{z, 0}^{A*}(\delta_{u})]} + \underset{\textcircled{2}\text{Immediate effect}}{\mathbb{E}_{b^U} [\cV_{z, 0}^{A*}(\delta_u)] - \cV_{z, 0}^{A*}(b^U)}.
    \end{equation*}
\end{prop}

Proposition~\ref{prop:measure.advantage} decomposes the advantage of measuring into two components:
the immediate effect (\textcircled{2}), which arises from removing uncertainty in the current decision and is always nonnegative by Jensen's inequality;
and the delayed effect (\textcircled{1}), which reflects how measuring affects the distribution of the next-step state and may negatively impact future rewards.

In the special case where $\cV_{z, 1}^{A*} = \cV_{z, 0}^{A*}$ (that is, measuring does not affect the environment), the delayed effect vanishes and the advantage is always nonnegative. In this regime, measuring strictly improves the policy value.

\begin{figure*}
    \centering
    \centering
    \begin{subfigure}{0.23\textwidth}
        \centering
        \includegraphics[width=\textwidth]{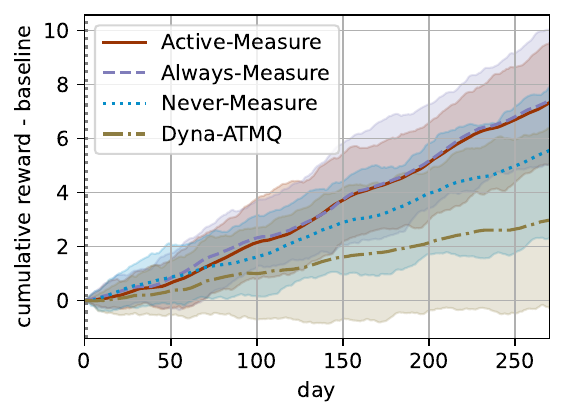}
        \caption{Minimal +, zero -}
        \label{subfig:s1.r}
    \end{subfigure}
    \hfill
    \begin{subfigure}{0.23\textwidth}
        \centering
        \includegraphics[width=\textwidth]{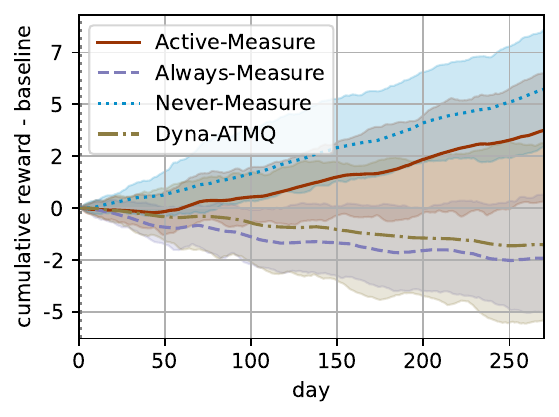}
        \caption{Minimal +, minimal -}
        \label{subfig:s4.r}
    \end{subfigure}
    \hfill
    \begin{subfigure}{0.23\textwidth}
        \centering
        \includegraphics[width=\textwidth]{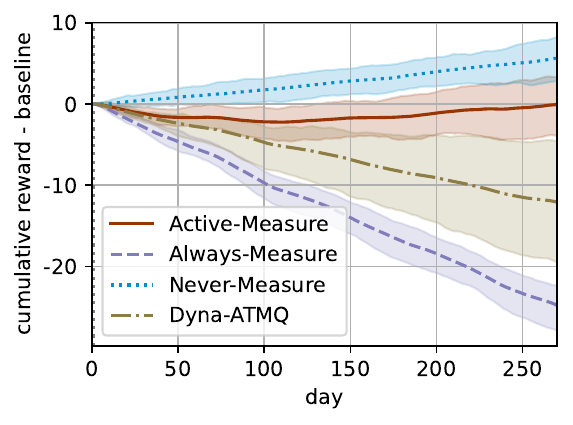}
        \caption{Minimal +, small -}
        \label{subfig:s7.r}
    \end{subfigure}
    \hfill
    \begin{subfigure}{0.23\textwidth}
        \centering
        \includegraphics[width=\textwidth]{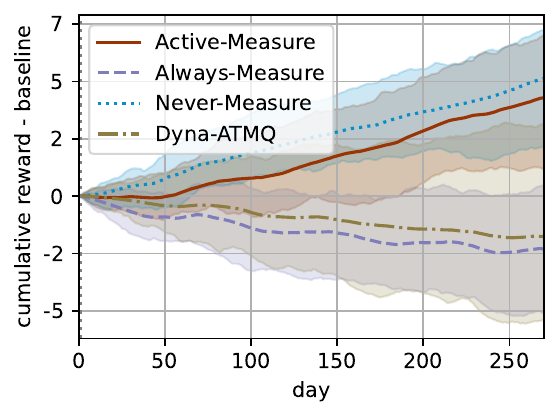}
        \caption{Minimal +, zero $O_t$}
        \label{subfig:s10.r}
    \end{subfigure}
    \hfill
    \begin{subfigure}{0.23\textwidth}
        \centering
        \includegraphics[width=\textwidth]{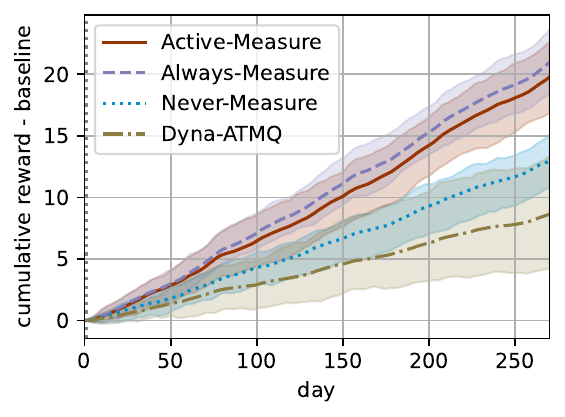}
        \caption{Small +, zero -}
        \label{subfig:s2.r}
    \end{subfigure}
    \hfill
    \begin{subfigure}{0.23\textwidth}
        \centering
        \includegraphics[width=\textwidth]{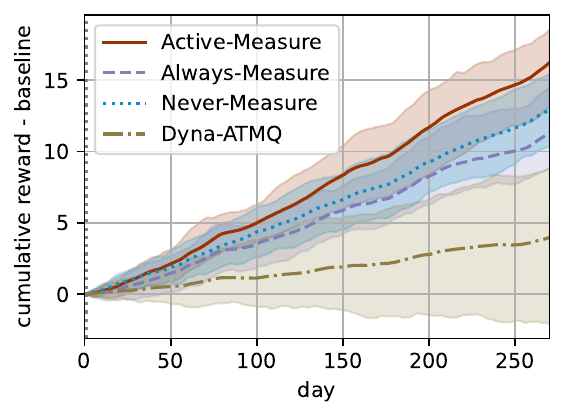}
        \caption{Small +, minimal -}
        \label{subfig:s5.r}
    \end{subfigure}
    \hfill
    \begin{subfigure}{0.23\textwidth}
        \centering
        \includegraphics[width=\textwidth]{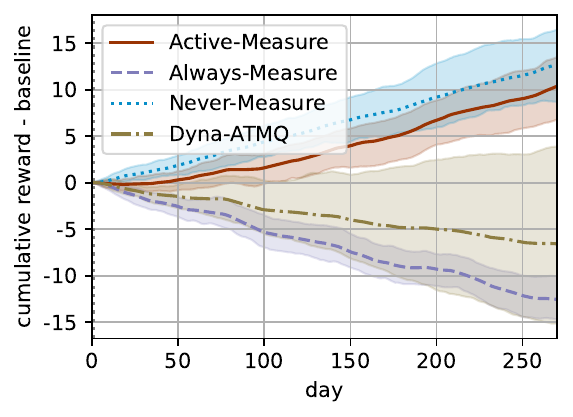}
        \caption{Small +, small -}
        \label{subfig:s8.r}
    \end{subfigure}
    \hfill
    \begin{subfigure}{0.23\textwidth}
        \centering
        \includegraphics[width=\textwidth]{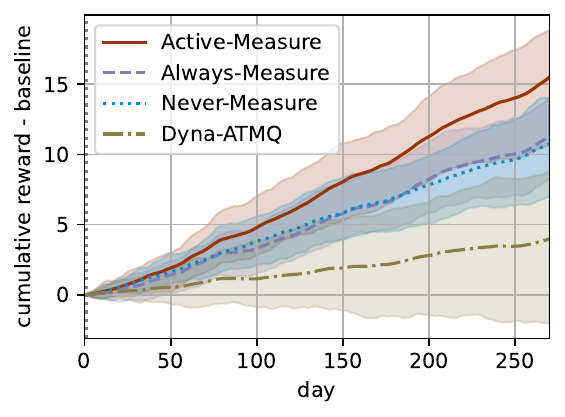}
        \caption{Small +, zero $O_t$}
        \label{subfig:s11.r}
    \end{subfigure}
    \hfill
    \begin{subfigure}{0.23\textwidth}
        \centering
        \includegraphics[width=\textwidth]{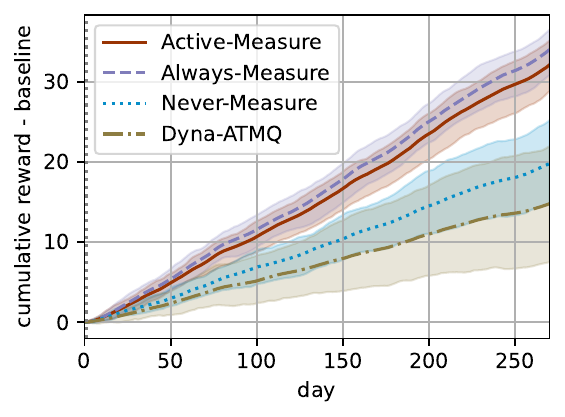}
        \caption{Medium +, zero -}
        \label{subfig:s3.r}
    \end{subfigure}
    \hfill
    \begin{subfigure}{0.23\textwidth}
        \centering
        \includegraphics[width=\textwidth]{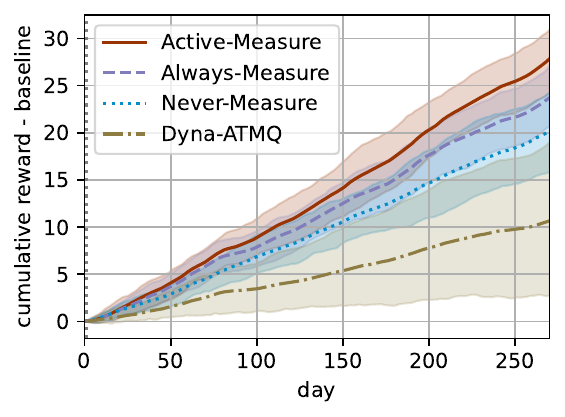}
        \caption{Medium +, minimal -}
        \label{subfig:s6.r}
    \end{subfigure}
    \hfill
    \begin{subfigure}{0.23\textwidth}
        \centering
        \includegraphics[width=\textwidth]{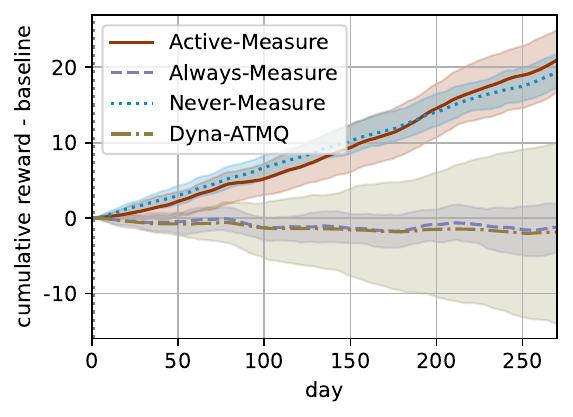}
        \caption{Medium +, small -}
        \label{subfig:s9.r}
    \end{subfigure}
    \hfill
    \begin{subfigure}{0.23\textwidth}
        \centering
        \includegraphics[width=\textwidth]{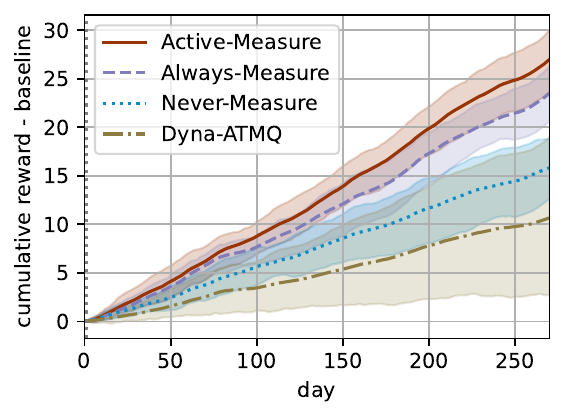}
        \caption{Medium +, zero $O_t$}
        \label{subfig:s12.r}
    \end{subfigure}
    \caption{The average cumulative reward, subtracting the average cumulative rewards of the zero policy.}
    \label{fig:simulation.results.r}
\end{figure*}

\section{Application} \label{sec:application}

We apply the proposed algorithm to HeartSteps, a digital intervention designed to help users increase and maintain physical activity (PA) levels.  
Figure~\ref{fig:dag} shows a simplified causal directed acyclic graph (DAG) for HeartSteps, which represents the strongest causal relations among the variables.  
Each time $t$ represents a day.  
The reward $R_t$ is the user's commitment to being active on day $t$.  
The control action $A_t$ is whether to send a walking-suggestion notification during day $t$.  
The measurement action $I_{t}$ is whether to send a survey to query the user about $R_{t-1}$.  
The emission $O_t$ can be the number of daily unprompted bouts of PA.  
The engagement $E_t$ captures the negative effects of the two actions.  
Excessive notifications or surveys reduce $E_t$ and thereby reduce the effectiveness of interventions $A_t$ on $R_t$.  
The context $C_t$ represents the evolving needs of the user, e.g., the logarithm of the prior 30-minute step count before the intervention time.  
The proximal outcome $M_t$ is a mediator between $A_t$ and $R_t$, e.g., the logarithm of the post 30-minute step count.  
Key factors that affect long-term behavior change are $R_t$ and $E_t$.
We utilize the public simulation testbed developed by \citet{gao2025harnessing}.  
It contains 42 different environments, constructed from the data on each of the 42 participants in HeartSteps.  
We can show that Figure~\ref{fig:dag} is a special case of an AOMDP (see Appendix~\ref{sec:heartsteps.aomdp}).  
Implementation details are provided in Appendices~\ref{sec:posterior.parameters}-\ref{sec:hyperparameters}.  

We compare the proposed active-measure algorithm with the always-measure and never-measure algorithms, which always take $I_{t, 1}$ to be one or zero and choose $A_{t, 2}$ with RLSVI.  
It is also compared against Dyna-ATMQ \citep{krale2023act}, which focuses on discrete states and assumes a fixed measurement cost observed together with the reward. 
ATMQ learns the negative effect only from the observed cost.  
Therefore, when implementing ATMQ, we discretize the states and treat the cost as a tuning parameter.  
The details of the baseline algorithms are provided in Appendices~\ref{sec:always.never.measure} and~\ref{sec:dyna.atmq}.

We consider different scenarios and report the average cumulative reward $\sum_{t=1}^T R_t$ across 42 users (environments), subtracting the average cumulative reward of the zero policy that takes $I_{t, 1} = A_{t, 2} = 0$ for all $t$.  
The experiment is repeated 50 times for each method in each scenario.  
Figure~\ref{fig:simulation.results.r} shows the average and 95\% confidence intervals across the 50 replications for the average cumulative reward.  
The first row represents the scenarios with the minimal positive effect of the interaction $A_{t, 2} R_{t - 1}$ on the next reward $R_t$. 
The second and third rows correspond to scenarios with small and medium positive effect sizes of $A_{t, 2} R_{t - 1}$ on $R_t$.
The first column represents the scenarios with no negative effect of the measurement action $I_{t, 1}$ on the next reward $R_t$.  
The second and third columns correspond to scenarios with minimal and small negative effect sizes of $I_{t, 1}$ on $R_t$.
The last column has the same settings as the second column but sets the strength of $R_t \to O_t$ to zero, i.e., the passively collected emission $O_t$ no longer help infer $R_t$.  
Details of testbed variants and their effect sizes are provided in Appendix~\ref{sec:testbed}.

When there is no negative effect of the measurement action in Subfigures~\eqref{subfig:s1.r}, \eqref{subfig:s2.r}, and~\eqref{subfig:s3.r}, the always-measure algorithm performs the best as it does not need to learn to measure, while the active-measure algorithm follows closely.  
When the measurement action has a minimal negative effect on the reward in Subfigures~\eqref{subfig:s7.r}, \eqref{subfig:s8.r}, and~\eqref{subfig:s9.r}, the never-measure algorithm performs best as it does not need to learn the negative effect.  
Active-measure has lower cumulative reward at the beginning, but it gradually picks up the negative signal while learning the transition and emission functions, and its cumulative reward increases faster than never-measure later on.  
When the measurement action has a small negative effect, with a small or medium positive effect of the control action in Subfigures~\eqref{subfig:s5.r} and~\eqref{subfig:s6.r}, active-measure performs significantly better than the baseline methods.  
Recall that the environment for subfigures~\eqref{subfig:s4.r} and~\eqref{subfig:s7.r} has a minimal effect of $A_{t, 2} R_{t - 1}$ on the reward.  
Under these scenarios, the observed states $E_{t - 1}$ and $C_{t, 2}$ may dominate the decision policy,  as there is no need to learn the latent state $R_{t - 1}$ perfectly. When a negative effect of $I_{t, 1}$ exists, never-measure performs better.  
An increased effect of $A_{t, 2} R_{t - 1}$ means that different values of the latent state $R_{t - 1}$ may flip the sign of the optimal action $A_{t, 2}$, and thus the benefit of measuring is more significant in Subfigures~\eqref{subfig:s3.r} and~\eqref{subfig:s6.r}.  
Comparing Subfigures~\eqref{subfig:s6.r} and~\eqref{subfig:s12.r}, we see that without an informative emission $O_t$, the cumulative reward of never-measure decreases significantly.  
Dyna-ATMQ generally achieves lower cumulative rewards since it requires discretizing the states and cannot detect small changes in the continuous rewards.  
See Appendix~\ref{sec:additional.simulation.results} for additional results. 

\section{Discussion}

The proposed algorithm can be naturally extended to problems with multiple measurement actions or control actions.  
For example, in HeartSteps, there can be two possible digital interventions per day.  
Such problems can still be fit into the periodic POMDP framework.  
For the $n$-step TD prediction used to construct the target of the control action in RLSVI, $n$ now depends on the time of the next nonzero instantaneous reward.  
In addition, while we focus on settings where the measurement action has no instantaneous reward, the proposed method can incorporate it in the target of the measurement action in RLSVI.


\subsubsection*{Acknowledgements}
This research was funded by NIH grants 2R01HL125440, P50DA054039, P41EB028242, UH3DE028723, U01CA229445, and 5P30AG073107-03 GY3 Pilots. Susan Murphy holds concurrent appointments at Harvard University and as an Amazon Scholar. This paper describes work performed at Harvard University and is not associated with Amazon.

\bibliography{bibfile}

\clearpage
\appendix
\thispagestyle{empty}

\counterwithin{thm}{section}
\counterwithin{figure}{section}
\counterwithin{table}{section}

\section{Definitions and Proofs} \label{sec:theoretical.proof}

In this section, we define periodic POMDPs and periodic belief MDPs and provide proofs of the theoretical results.

\subsection{Periodic POMDP and Periodic Belief MDP} \label{sec:periodic.belief.mdp}



For conciseness of notation, we use the index $(t, K + 1)$ to refer to the index $(t + 1, 1)$, and $(t, 0)$ to refer to $(t - 1, K)$.

\begin{dfn}[$K$-Periodic POMDP] \label{def:periodic.momdp}
    A \textit{$K$-periodic POMDP} is a tuple $(\cS_{1:K}, \allowbreak \cO_{1:K}, \allowbreak \cA_{1:K}, \allowbreak \bbT_{1:K}, \allowbreak \bbO_{1:K}, \allowbreak r_{1:K}, \allowbreak \gamma)$, where
    $\cS_k \subset \bbR^{d_{S_k}}$ is the latent state space,
    $\cO_k \subset \bbR^{d_{O_k}}$ is the emission space for $\cS_k$,
    $\cA_k$ is the action space,
    and $\gamma$ is the discount factor.
    The transition function for the latent state is $\bbT_k: \cS_{k-1} \times \cA_{k-1} \mapsto \Delta (\cS_{k})$, with $\bbT_k (s_{t, k} \mid s_{t, k - 1}, a_{t, k - 1})$ being the p.d.f. of $S_{t, k}$ given $S_{t, k - 1} = s_{t, k - 1}$ and $A_{t, k - 1} = a_{t, k - 1}$.
    The emission function is $\bbO_k: \cS_{k} \mapsto \Delta (\cO_k)$, with $\bbO_k (o_{t, k} \mid s_{t, k})$ being the p.d.f. of $O_{t, k}$ given $S_{t, k} = s_{t, k}$.
    The reward function $r_k: \cS_{k+1} \mapsto \bbR$ is a known deterministic function of the next latent state.
\end{dfn}

\begin{dfn}[$K$-Periodic Belief MDP] \label{def:periodic.belief.mdp}
    A \textit{$K$-periodic belief MDP} is a tuple $(\cB_{1:K}, \cA_{1:K}, \bbT_{1:K}, r_{1:K}, \gamma)$, where $\cB_k = \Delta (\cS_k)$ is the set of belief states over the latent state space $\cS_k$ at time $k$ in a period, $\cA_k$ is the action space, and $\gamma$ is the discount factor. 
    The transition function for the belief states is $\bbT_k: \cB_{k-1} \times \cA_{k-1} \mapsto \cB_k$. 
    With $r_k: \cS_{k+1} \mapsto \bbR$ being a known reward function of the next latent state, the reward of the belief state is $r_k (b_{t, k + 1}^S) = \int r_k (s) b_{t, k + 1}^S (s) ds$.
\end{dfn}

A sequence of Markov stationary policies is denoted by $\bpi := \{\pi_{1:K}\}$, where $\pi_k: \cB_k \mapsto \cA_k$.  
The Q-function of a periodic belief MDP under policy $\bpi$ is
\begin{equation*} \label{equ:definition.Q}
    \cQ_k^{\bpi} (b^S_{t, k}, a_{t, k}) := 
    \bbE^{\bpi} \brce{
    \sum_{(i, j): (i, j) \ge (t, k)} 
    \gamma^{K (i - t) + j - k} \cdot r_{j} (b^S_{i, j + 1})
    \,\middle|\, b^S_{t, k}, A_{t, k} = a_{t, k} },
\end{equation*}
where $(i, j) > (t, k)$ means $i = t$ and $j > k$, or $i > t$, and $(i, j) = (t, k)$ means $i = t$ and $j = k$.  
The value function is defined as $\cV_k (b^S_{t, k}) = \bbE^{\bpi} \{ \cQ_k (b^S_{t, k}, A_{t, k}) \mid b^S_{t, k} \}$.  
Extending the results from the periodic MDP to the periodic belief MDP, the Bellman optimality equation for the belief state is
\begin{equation} \label{equ:bellman.optimality}
    \begin{split}
        \cQ^*_k (b^S_{t, k}, a_{t, k}) = 
        \bbE \brce{ r_k (b^S_{t, k + 1}) 
        + \gamma_k \max_{a_{t, k + 1} \in \cA_{k + 1}} 
        \cQ^*_{k + 1} (b^S_{t, k + 1}, a_{t, k + 1}) 
        \,\middle|\, b^S_{t, k}, A_{t, k} = a_{t, k} }
    \end{split}
\end{equation}
for $k \in \{1:K\}$.  
The optimal value function is then $\cV^*_k (b^S_{t, k}) = \max_{a_{t, k} \in \cA_k} \cQ^*_k (b^S_{t, k}, a_{t, k})$.  


Theoretically, the belief state can be updated as  
$b^S_{t, k + 1} (s_{t, k + 1}) = \bbP (s_{t, k + 1}, o_{t, k + 1} \mid b^S_{t, k}, a_{t, k}) / \bbP (o_{t, k + 1} \mid b^S_{t, k}, a_{t, k})$,  
where  
$\bbP (s_{t, k + 1}, o_{t, k + 1} \mid b^S_{t, k}, a_{t, k})  
= \int_{\cS_{k}} \bbT_{k + 1} (s_{t, k + 1} \mid s, a_{t, k}) \bbO_{k + 1} (o_{t, k + 1} \mid s_{t, k + 1}) b^S_{t, k} (s) \, d s$,  
and  
$\bbP (o_{t, k + 1} \mid b^S_{t, k}, a_{t, k})  
= \int_{\cS_{k + 1}} \bbP (s', o_{t, k + 1} \mid b^S_{t, k}, a_{t, k}) \, d s'$.  

Note that the right-hand side of Equation~\eqref{equ:bellman.optimality} is an expectation over $O_{t, k + 1}$, i.e.,
\begin{equation*}
    \int \brck{r_k (b^S_{t, k + 1}) + \gamma_k \max_{a_{t, k + 1} \in \cA_{k + 1}} \cQ^*_{k + 1} (b^S_{t, k + 1}, a_{t, k + 1})} 
    p (o_{t, k + 1} \mid b^S_{t, k}, a_{t, k}) \, d o_{t, k + 1},
\end{equation*}
where $b^S_{t, k + 1}$ is a function of $a_{t, k}$, $o_{t, k + 1}$, and $b^S_{t, k}$.  
That is, $b^S_{t, k + 1} (s_{t, k + 1}) = p (s_{t, k + 1} \mid b^S_{t, k}, a_{t, k}, o_{t, k + 1})$.  
The maximization is taken separately for each $o_{t, k + 1}$.

\subsection{AOMDP as a Periodic POMDP}

In this subsection, we prove Lemma~\ref{lem:aomdp.periodic.pomdp}, which formulates the AOMDP as a 2-periodic POMDP.  

\begin{proof}

With some overload of notation, recall that the components $(\cZ, \cU, \cO, \cA, \cI, \bbT, \bbO, r, \gamma)$ of an AOMDP are not indexed, whereas the components $(\cS_{1:K}, \cO_{1:K}, \cA_{1:K}, \bbT_{1:K}, \bbO_{1:K}, r_{1:K}, \gamma_{1:K})$ of a $K$-periodic POMDP are indexed by the time step $k$.  
In an AOMDP, the variables $Z_t, U_t, O_t$ are indexed only by the period number $t$, with the actions $I_{t, 1}$ and $A_{t, 2}$ indexed by both $t$ and $k$.  
In contrast, in a $K$-periodic POMDP, the variables $S_{t, k}, O_{t, k}, A_{t, k}$ are all indexed by both $t$ and $k$.

\paragraph{Action space.}
In an AOMDP, there are two decision times in each period $t$.  
The measure action $I_{t, 1}$ and the control action $A_{t, 2}$ are the two actions in a period.  
The action spaces in the 2-periodic POMDP are $\cA_1 = \cI = \{0, 1\}$ and $\cA_2 = \cA = \{0, 1\}$.

\paragraph{State space.}
Define the latent state for $I_{t, 1}$ as $S^I_{t, 1} = [Z_{t}, U_{t}]$, and the latent state for $A_{t, 2}$ as $S^A_{t, 2} = [Z_{t}, U_{t}, I_{t, 1}]$.  
The state spaces in the 2-periodic POMDP are $\cS_1 = \bbR^{d_Z + d_U}$ and $\cS_2 = \bbR^{d_Z + d_U} \times \{0, 1\}$.

\paragraph{Emission space.}
Since $O_t \in \cO$ and $I_{t, 1} U_t \in \bbR$ is the emission of $U_t$ in an AOMDP, the emissions in the 2-periodic POMDP can be defined as  
$O^I_{t, 1} = [Z_{t}, O_{t}]$ and $O^A_{t, 2} = [Z_{t}, O_{t}, I_{t, 1} U_{t}, I_{t, 1}]$.  
The corresponding emission spaces are $\cO_1 = \bbR^{d_Z} \times \cO$ and $\cO_2 = \bbR^{d_Z} \times \cO \times \bbR^{d_U} \times \cI$.

\paragraph{Discount factor.}
Since the instantaneous reward of $I_{t, 1}$ is zero and the discount factor on $R_t = r(Z_{t + 1}, U_{t + 1})$ is $\gamma$, we define $\gamma_1 = \gamma_2 = \sqrt{\gamma}$.  

\paragraph{Transition function.}
For the state $S^A_{t - 1, 2} = s^A_{t - 1, 2} := [z_{t - 1}, u_{t - 1}, i_{t - 1, 1}]$, the action $A_{t - 1, 2} = a_{t - 1, 2}$, and the next state $S^I_{t, 1} = s^I := [z, u]$,  
the transition function at time $k = 1$ is defined as  
\[
\bbT_1 (s^I \mid s^A_{t - 1, 2}, a_{t - 1, 2}) = \bbT (z, u \mid z_{t - 1}, u_{t - 1}, i_{t - 1, 1}, a_{t - 1, 2}).
\]  

For the state $S^I_{t, 1} = s^I_{t, 1} := [z_t, u_t]$, the action $I_{t, 1} = i_{t, 1}$, and the next state $S^A_{t, 2} = s^A := [z, u, i]$,  
the transition function at time $k = 2$ is defined as  
\[
\bbT_2 (s^A \mid s^I_{t, 1}, i_{t, 1}) = \delta (z - z_t) \, \delta (u - u_t) \, \bbone(i = i_t).
\]

\paragraph{Emission function.}
For the state $S^I_{t, 1} = s^I_{t, 1} := [z_{t}, u_{t}]$ and emission $O^I_{t, 1} = o^I := [z, o]$, the emission function is  
\[
\bbO_1 (o^I \mid s^I_{t, 1}) = \delta (z - z_t) \, \bbO (o \mid u_t).
\]  
Similarly, for the state $S^A_{t, 2} = s^A_{t, 2} := [z_{t}, u_{t}, i_{t, 1}]$ and emission $O^A_{t, 2} = o^A := [z, o, u, i]$, the emission function is  
\[
\bbO_2 (o^A \mid s^A_{t, 2}) = 
\delta (z - z_t) \, \bbO (o \mid u_t) \,
\big[ \bbone (i_{t, 1} = 1) \delta (u - u_t) + \bbone (i_{t, 1} = 0) \delta (u - 0) \big] \,
\bbone (i = i_{t, 1}).
\]  

\paragraph{Reward function.}
Since the instantaneous reward of $I_{t, 1}$ is zero, the reward function at time $k = 1$ is  
\[
r_1 (s^A_{t, 2}) = 0 \quad \text{for all } s^A_{t, 2} \in \cS_2.
\]  
The reward function at time $k = 2$ is  
\[
r_2 (s^I_{t + 1, 1}) = r (z_{t + 1}, u_{t + 1}) \quad 
\text{for } s^I_{t + 1, 1} := [z_{t + 1}, u_{t + 1}].
\]  
\end{proof}

\subsection{Sample Complexity Improvement}

In this subsection, we prove Proposition~\ref{prop:AOMDP.alpha.revealing}.

\begin{proof}

    To utilize the conclusion in \citet{liu2022partially}, we redefine the action as $\widetilde{A}_t = [I_{t, 1}, A_{t, 2}]$, the state as $\widetilde{S}_t = [Z_{t - 1}, U_{t - 1}, I_{t - 1, 1}, A_{t - 1, 2}, Z_t, U_t]$, and the emission as $\widetilde{O}_t = [Z_{t - 1}, O_{t - 1}, I_{t - 1, 1} U_{t - 1}, I_{t - 1, 1}, A_{t - 1, 2}, Z_t, O_t]$, so that $\widetilde{O}_t$ depends only on $\widetilde{S}_t$.  
    This can be viewed as a special case of solving an AOMDP as a periodic POMDP.  
    
    For the measurement action $I_{t, 1}$, the state $\widetilde{S}_t$ is equivalent to $S^I_{t, 1} = [Z_{t}, U_{t}]$, since  
    $Z_{t - 1}, U_{t - 1}, I_{t - 1, 1}, A_{t - 1, 2} \indep Z_{t + 1}, U_{t + 1} \mid b^{S^I}_{t, 1}, I_{t, 1}$.  
    For the control action $A_{t, 2}$, the state $\widetilde{S}_t$ is a special case of $S^A_{t, 2} = [Z_{t}, U_{t}, I_{t, 1}]$, since  
    $Z_{t - 1}, U_{t - 1}, I_{t - 1, 1}, A_{t - 1, 2} \indep Z_{t + 1}, U_{t + 1} \mid b^{S^I}_{t, 1}, I_{t, 1}$ and $\{Z_t, U_t\} \subsetneq S^A_{t, 2}$.  
    Selecting $A_{t, 2}$ based on $\widetilde{S}_t$ therefore provides a lower bound for the sample complexity of selecting $A_{t, 2}$ based on $S^A_{t, 2}$.  
    
    Now we investigate the submatrix $M_{t, \widetilde{a}}$, corresponding to the rows of $M_{t}$ where the action is $\widetilde{a}$, i.e., $(M_{t, \widetilde{a}})_{o, s} = [M_{t}]_{(\widetilde{a}, o), s}$.  
    Suppose the sizes of the observed state, latent state, emission, measurement action, and control action are $Z$, $U$, $O$, $I$, and $A$, respectively.  
    Then the size of the state space is $\widetilde{S} = Z \times U \times I \times A \times Z \times U$, and the size of the emission space is $\widetilde{O} = Z \times O \times U \times I \times A \times Z \times O$.  
    Thus, the submatrix $M_{t, \widetilde{a}}$ has dimension $\widetilde{O}^2 \times \widetilde{S}$.  
    
    When we take $\widetilde{a} = [1, a]$ for any $a \in \cA$ (i.e., $I_{t, 1} = 1$), we will show that $M_{t, \widetilde{a}}$ contains a square submatrix of size $\widetilde{S} \times \widetilde{S}$ that has full rank.  
    Specifically, denote the row value as $[z_0, o_0, iu_0, i_0, a_0, z_1, o_1, z_1, o_1, iu_1, i_1, a_1, z_2, o_2]$ (the concatenation of emissions $\widetilde{O}_t$ and $\widetilde{O}_{t + 1}$), and the column value as $[z_0', u_0', i_0', a_0', z_1', u_1']$ (the state $\widetilde{S}_t$).  
    Consider the submatrix obtained by fixing values of $o_0$, $o_1$, $o_2$, and $z_2$.  
    Then the size of this submatrix is $\widetilde{S} \times \widetilde{S}$, as only $z_0, iu_0, i_0, a_0, z_1$, and $iu_1$ are allowed to vary.  
    This submatrix is diagonal, with diagonal entries equal to one for indices $z_0 = z_0'$, $iu_0 = u_0'$, $i_0 = i_0'$, $a_0 = a_0'$, $z_1 = z_1'$, and $iu_1 = u_1'$.  
    All off-diagonal entries are zero.  
    Therefore, $\sigma_{\widetilde{S}} (M_{t, \widetilde{a}}) = 1$.  
    
    According to Proposition~3 of \citet{liu2022partially}, since $\max_{i} \sigma_{\widetilde{S}} (M_{t, \widetilde{a}}) \ge 1$ with the maximizer $\widetilde{a} = [1, a]$, the 2-step 1-weakly revealing condition is satisfied.  
\end{proof}

\subsection{Bellman Optimality Equation for the Measure Action} \label{sec:target.I}

In this subsection, we prove for Proposition~\ref{prop:measure.advantage}.

\begin{proof}
    
To derive the target for the measure action $I$ in RLSVI, first note that the emission distribution of $I_{t, 1} U_{t}$ is  
\begin{equation*}
\begin{split}
    p_{I_{t, 1} U_{t}} (u_t \mid b^{U}_{t, 1}, I_{t, 1}) = 
    \begin{cases}
        b^{U}_{t, 1} (u_t), & \text{if } I_{t, 1} = 1, \\
        \delta (u_t - 0), & \text{if } I_{t, 1} = 0,
    \end{cases}
\end{split}
\end{equation*}
where $\delta(\cdot)$ is the Dirac delta function.  

When $I_{t, 1} = 1$, the p.d.f. of the belief state $b^{S^A}_{t, 2}$ at value $s^A = [z, u, i]$ given the previous belief state $b^{S^I}_{t, 1} = \delta_{z_{t}} \otimes b^U_{t, 1}$, the last action $I_{t, 1} = 1$, and the emission $I_{t, 1} U_{t} = u_{t}$ is
\begin{equation*}
\begin{split}
    & b^{S^A}_{t, 2} (s^A) 
    = p (s^A \mid b^{S^I}_{t, 1}, I_{t, 1} = 1, I_{t, 1} U_{t} = u_{t}) \\
    = & p_{Z_t} (z \mid b^{S^I}_{t, 1}, I_{t, 1} = 1, I_{t, 1} U_{t} = u_{t})
    \cdot P (I_{t, 1} = i \mid b^{S^I}_{t, 1}, I_{t, 1} = 1, I_{t, 1} U_{t} = u_{t}, Z_t = z) \\ 
    & \cdot p_{U_t} (u \mid \delta_{z_{t}} \otimes b^U_{t, 1}, I_{t, 1} = 1, I_{t, 1} U_{t} = u_{t}, Z_t = z, I_{t, 1} = i) \\
    = & p_{Z_t} (z \mid \delta_{z_{t}})
    \cdot P (I_{t, 1} = i \mid I_{t, 1} = 1) 
    \cdot \frac{p_{U_t} (u \mid b^U_{t, 1}, I_{t, 1} = 1) \, p_{I_{t, 1} U_t} (u_t \mid U_t = u, I_{t, 1} = 1)}{p_{I_{t, 1} U_t} (u_t \mid b^U_{t, 1}, I_{t, 1} = 1)} \\
    = & \delta (z - z_{t}) \, \delta (u - u_{t}) \, \bbone (i = 1).
\end{split}
\end{equation*}
The last equality holds since $p_{I_{t, 1} U_t} (u_t \mid U_t = u, I_{t, 1} = 1) = \delta (u - u_{t})$.  
Therefore, the transition function from $b^{S^I}_{t, 1}$ to $b^{S^A}_{t, 2}$ is  
\begin{equation*}
\begin{split}
    \cT (b^{S^A}_{t, 2} \mid b^{S^I}_{t, 1}, I_{t, 1} = 1) 
    = & \int \delta_{\delta_{z_t} \otimes \delta_{u_t} \otimes \delta_{1}} \,
    p_{I_{t, 1} U_{t}} (u_t \mid b^{U}_{t, 1}, I_{t, 1}) \, d u_t \\
    = & \int \delta_{\delta_{z_t} \otimes \delta_{u_t} \otimes \delta_{1}} \,
    b^{U}_{t, 1} (u_t) \, d u_t,
\end{split}
\end{equation*}
since $p_{I_{t, 1} U_{t}} (u_t \mid b^{U}_{t, 1}, I_{t, 1}) = b^{U}_{t, 1} (u_t)$ when $I_{t, 1} = 1$.  
Then, based on the Bellman optimality equation~\eqref{equ:bellman.optimality.I},  
\begin{equation} \label{equ:Q.I1}
\begin{split}
    \cQ^{I*} (b^{S^I}_{t, 1}, I_{t, 1})
    = & \int \brck{\sqrt{\gamma} \max_{a_{t, 2} \in \cA} \cQ^{A*} (b^{S^A}_{t, 2}, a_{t, 2})} \,
    p_{I_{t, 1} U_{t}} (u_t \mid b^{U}_{t, 1}, I_{t, 1}) \, d u_t \\
    = & \sqrt{\gamma} \int \max_{a_{t, 2} \in \cA} 
    Q^{A*} ( \delta_{z_t} \otimes \delta_{u_t} \otimes \delta_{1}, a_{t, 2} ) \,
    b^{U}_{t, 1} (u_t) \, d u_t,
\end{split}
\end{equation}
since the instantaneous reward of $I_{t, 1}$ is zero.

On the other hand, when $I_{t, 1} = 0$,  
\begin{equation*}
\begin{split}
    & b^{S^A}_{t, 2} (s^A) 
    = p (s^A \mid b^{S^I}_{t, 1}, I_{t, 1} = 0, I_{t, 1} U_{t} = u_t) \\
    = & p_{Z_t} (z \mid \delta_{z_{t}})
    \cdot P (I_{t, 1} = i \mid I_{t, 1} = 0) 
    \cdot \frac{p_{U_t} (u \mid b^U_{t, 1}, I_{t, 1} = 0) \, p_{I_{t, 1} U_t} (u_t \mid U_t = u, I_{t, 1} = 0)}{p_{I_{t, 1} U_t} (u_t \mid b^U_{t, 1}, I_{t, 1} = 0)} \\
    = & \delta (z - z_{t}) \, b^{U}_{t, 1} (u) \, \bbone (i = 0),
\end{split}
\end{equation*}
The last equality holds since $p_{I_{t, 1} U_t} (u_t \mid U_t = u, I_{t, 1} = 0) = p_{I_{t, 1} U_t} (u_t \mid b^U_{t, 1}, I_{t, 1} = 0) = \delta (u_t - 0)$  
and $p_{U_t} (u \mid b^U_{t, 1}, I_{t, 1} = 0) = b^U_{t, 1} (u)$.  
Therefore, the transition function from $b^{S^I}_{t, 1}$ to $b^{S^A}_{t, 2}$ is  
\begin{equation*}
\begin{split}
    \cT (b^{S^A}_{t, 2} \mid b^{S^I}_{t, 1}, I_{t, 1}) 
    = & \int \delta_{\delta_{z_t} \otimes b^{U}_{t, 1} \otimes \delta_{0}} \,
    p_{I_{t, 1} U_{t}} (u_t \mid b^{U}_{t, 1}, I_{t, 1}) \, d u_t \\
    = & \delta_{\delta_{z_t} \otimes b^{U}_{t, 1} \otimes \delta_{0}},
\end{split}
\end{equation*}
since $p_{I_{t, 1} U_{t}} (u_t \mid b^{U}_{t, 1}, I_{t, 1}) = \delta (u_t - 0)$ when $I_{t, 1} = 0$.  
Similarly, based on the Bellman optimality equation~\eqref{equ:bellman.optimality},  
\begin{equation} \label{equ:Q.I0}
\begin{split}
    \cQ^{I*} (b^{S^I}_{t, 1}, I_{t, 1})
    = & \int \brck{\sqrt{\gamma} \max_{a_{t, 2} \in \cA} \cQ^{A*} (b^{S^A}_{t, 2}, a_{t, 2})} \,
    p_{I_{t, 1} U_{t}} (u_t \mid b^{U}_{t, 1}, I_{t, 1}) \, d u_t \\
    = & \sqrt{\gamma} \max_{a_{t, 2} \in \cA} 
    Q^{A*} ( \delta_{z_t} \otimes b^{U}_{t, 1} \otimes \delta_{0}, a_{t, 2} ).
\end{split}
\end{equation}

Comparing equations \eqref{equ:Q.I1} and \eqref{equ:Q.I0},  
we see that in \eqref{equ:Q.I1} the optimal control action $a_{t, 2}$ is selected based on each possible value of the emission $I_{t, 1} U_{t}$,  
while in \eqref{equ:Q.I0} it is selected based on the entire belief state $b^{S^A}_{t, 2}$.  
Further, we have
\begin{equation*}
\begin{split}
    \cQ^{I*} (b^{S^I}_{t, 1}, 1) - \cQ^{I*} (b^{S^I}_{t, 1}, 0)
    = & \int \max_{a_{t, 2} \in \cA} Q^{A*} ( \delta_{z_t} \otimes \delta_{u_t}, a_{t, 2} ) \, b^{U}_{t, 1} (u_t) \, d u_t \\
    & \quad - \max_{a_{t, 2} \in \cA} Q^{A*} ( \delta_{z_t} \otimes b^{U}_{t, 1}, a_{t, 2} ) \\
    = & \, \bbE [\cV_1^{A*} (\delta_u)] - \cV_0^{A*} (b^{U}_{t, 1}) \\
    = & \, \bbE [\cV_1^{A*} (\delta_u)] - \bbE [\cV_0^{A*} (\delta_u)] 
        + \bbE [\cV_0^{A*} (\delta_u)] - \cV_0^{A*} (b^{U}_{t, 1}).
\end{split}
\end{equation*}
\end{proof}

\begin{rmk}
    When the measure action does not affect the environment, $I_{t, 1}$ is not part of the state $S^A_{t, 2}$.  
    In this case, we have $b^{S^A}_{t, 2} = \delta_{z_t} \otimes \delta_{u_t}$ when $I_{t, 1} = 1$ and $b^{S^A}_{t, 2} = \delta_{z_t} \otimes b^{U}_{t, 1}$ when $I_{t, 1} = 0$.  
    By Jensen's inequality, it follows that
    \begin{equation*}
        \cQ^{I*} (b^{S^I}_{t, 1}, 1)
        = \int \max_{a_{t, 2} \in \cA} Q^{A*} ( \delta_{z_t} \otimes \delta_{u_t}, a_{t, 2} ) \, b^{U}_{t, 1} (u_t) \, d u_t
        \ge \max_{a_{t, 2} \in \cA} Q^{A*} ( \delta_{z_t} \otimes b^{U}_{t, 1}, a_{t, 2} )
        = \cQ^{I*} (b^{S^I}_{t, 1}, 0).
    \end{equation*}
    However, this inequality may not hold when $I_{t, 1}$ does affect the environment.
\end{rmk}

\subsection{Belief Propagation With Sequential Monte Carlo} \label{sec:proof.belief.state}

\paragraph{Belief state $b_{t, 1}^U$.}
To obtain the belief state $b_{t, 1}^U$ for $U_{t}$, note that the joint posterior distribution of $U_{1:t}$ and $\btheta$ given the observed history $H_{t, 1} = \{Z_{1}, O_{1}, I_{1, 1}, I_{1, 1} U_{1}, A_{1, 2}, \dots, Z_{t}, O_{t}\}$ is
\begin{equation*}
    \begin{split}
        p ( U_{1:t}, \btheta \mid H_{t, 1} )
        = & p ( U_{1:t}, \btheta \mid H_{t - 1, 2}, A_{t - 1, 2}, Z_{t}, O_{t}) \\
        = & p ( U_{1:t-1} \mid H_{t - 1, 2}, A_{t - 1, 2} ) \cdot \\
        & p ( \btheta \mid U_{1:t-1}, H_{t - 1, 2}, A_{t - 1, 2} ) \cdot \\
        & p ( Z_{t} \mid \btheta, U_{1:t-1}, H_{t - 1, 2}, A_{t - 1, 2} ) \cdot \\
        & p ( U_{t} \mid Z_{t}, \btheta, U_{1:t-1}, H_{t - 1, 2}, A_{t - 1, 2} ) \cdot \\
        & p ( O_{t} \mid U_{t}, Z_{t}, \btheta, U_{1:t-1}, H_{t - 1, 2}, A_{t - 1, 2} ) \cdot \\
        & \brck{ p ( Z_{t}, O_{t} \mid H_{t - 1, 2}, A_{t - 1, 2} ) }^{-1} \\
        \propto & p ( U_{1:t-1} \mid H_{t - 1, 2} ) \cdot
        p ( \btheta \mid U_{1:t-1}, H_{t - 1, 2} ) \cdot \\
        & p ( Z_{t} \mid \btheta, U_{t-1}, Z_{t-1}, I_{t - 1, 1}, A_{t - 1, 2} ) \cdot \\
        & p ( U_{t} \mid Z_{t}, \btheta, U_{t-1}, Z_{t-1}, I_{t - 1, 1}, A_{t - 1, 2} ) \cdot 
        p ( O_{t} \mid \btheta, U_{t} ).
    \end{split}
\end{equation*}
Here, $p ( U_{1:t-1} \mid H_{t - 1, 2}, A_{t - 1, 2} ) = p ( U_{1:t-1} \mid H_{t - 1, 2} )$ since $A_{t - 1, 2}$ is chosen only based on $H_{t - 1, 2}$, i.e., $A_{t - 1, 2} \indep U_{1:t-1} \mid H_{t - 1, 2}$.
Suppose the proposal distribution can be factorized as
\begin{equation*}
    q (U_{1:t}, \btheta \mid H_{t, 1} )
    = q (U_{t} \mid \btheta, U_{1:t-1}, H_{t, 1} ) \,
      q (\btheta \mid U_{1:t-1}, H_{t-1, 2} ) \,
      q (U_{1:t-1} \mid H_{t-1, 2} ).
\end{equation*}
Then the importance weight can be written as
\begin{equation*}
    \begin{split}
        \widetilde{W}_{t, 1} 
        = & \frac{p (U_{1:t}, \btheta \mid H_{t, 1} )}{q (U_{1:t}, \btheta \mid H_{t, 1} )} \\
        \propto & \; p ( U_{1:t-1} \mid H_{t - 1, 2} ) \cdot
        p ( \btheta \mid U_{1:t-1}, H_{t - 1, 2} ) \cdot 
        p ( Z_{t} \mid \btheta, U_{t-1}, Z_{t-1}, I_{t - 1, 1}, A_{t - 1, 2} ) \cdot \\
        & p ( U_{t} \mid Z_{t}, \btheta, U_{t-1}, Z_{t-1}, I_{t - 1, 1}, A_{t - 1, 2} ) \cdot 
        p ( O_{t} \mid \btheta, U_{t} ) \cdot \\
        & \brck{
        q (U_{1:t-1} \mid H_{t-1, 2} ) \,
        q (\btheta \mid U_{1:t-1}, H_{t-1, 2} ) \,
        q (U_{t} \mid \btheta, U_{1:t-1}, H_{t, 1} ) 
        }^{-1} \\
        = & \; W_{t-1, 2} \cdot
        \frac{p ( \btheta \mid U_{1:t-1}, H_{t - 1, 2} )}{q(\btheta \mid U_{1:t-1}, H_{t-1, 2})} \cdot
        \frac{p ( U_{t} \mid Z_{t}, \btheta, U_{t-1}, Z_{t-1}, I_{t - 1, 1}, A_{t - 1, 2} )}
             {q (U_{t} \mid \btheta, U_{1:t-1}, H_{t, 1} )} \cdot \\
        & p ( Z_{t} \mid \btheta, U_{t-1}, Z_{t-1}, I_{t - 1, 1}, A_{t - 1, 2} ) \cdot 
        p ( O_{t} \mid \btheta, U_{t} ),
    \end{split}
\end{equation*}
where 
\[
W_{t-1, 2} = \frac{p ( U_{1:t-1} \mid H_{t - 1, 2} )}{q (U_{1:t-1} \mid H_{t-1, 2} )}
\]
is the weight for the marginal posterior distribution of $U_{1:t-1}$ given the history $H_{t - 1, 2}$.
If we take 
\begin{align*}
    q (\btheta \mid U_{1:t-1}, H_{t-1, 2} ) &= p ( \btheta \mid U_{1:t-1}, H_{t - 1, 2} ), \\
    q (U_{t} \mid \btheta, U_{1:t-1}, H_{t, 1} ) &= p ( U_{t} \mid Z_{t}, \btheta, U_{t-1}, Z_{t-1}, I_{t - 1, 1}, A_{t - 1, 2} ),
\end{align*}
then the importance weight $\widetilde{W}_{t, 1}$ can be updated as
\begin{equation*}
    \widetilde{W}_{t, 1} \propto W_{t-1, 2} \cdot 
    p ( Z_{t} \mid \btheta, U_{t-1}, Z_{t-1}, I_{t - 1, 1}, A_{t - 1, 2} ) \cdot 
    p ( O_{t} \mid \btheta, U_{t} ).
\end{equation*}

\paragraph{Belief state $b_{t, 2}^U$.}
To obtain the belief state $b_{t, 1}^U$ for $U_{t}$, we break down the posterior distribution $p (U_{1:t}, \btheta \mid H_{t, 2} )$ separately for the cases $I_{t} = 0$ and $I_{t} = 1$.

When $I_{t} = 0$, the joint posterior distribution of $U_{1:t}$ and $\btheta$ given the observed history $H_{t, 2} = H_{t, 1} \cup \{I_{t, 1}, I_{t, 1} U_{t}\}$ is
\begin{equation*}
    \begin{split}
        p (U_{1:t}, \btheta \mid H_{t, 2} )
        = & p ( U_{1:t}, \btheta \mid H_{t, 1}, I_{t, 1}, I_{t, 1} U_{t} ) \\
        = & p ( U_{1:t}, \btheta \mid H_{t, 1}, I_{t, 1} ) \cdot \\
        & p ( I_{t, 1} U_{t} \mid \btheta, U_{1:t}, H_{t, 1}, I_{t, 1} ) \\
        & \brck{ p ( I_{t, 1} U_{t} \mid H_{t, 1}, I_{t, 1} ) }^{-1} \\
        \propto & p ( U_{1:t}, \btheta \mid H_{t, 1} ) \cdot 
        p ( I_{t, 1} U_{t} \mid U_{t}, I_{t, 1} ).
    \end{split}
\end{equation*}
Here, $p ( U_{1:t}, \btheta \mid H_{t, 1}, I_{t, 1} ) = p ( U_{1:t}, \btheta \mid H_{t, 1} )$ since $I_{t, 1}$ is chosen only based on $H_{t, 1}$, i.e., $I_{t, 1} \indep (U_{1:t}, \btheta) \mid H_{t, 1}$.
Take the proposal distribution as
\begin{equation*}
    q (U_{1:t}, \btheta \mid H_{t, 2} )
    = q (U_{1:t}, \btheta \mid H_{t, 1} ),
\end{equation*}
which is the same as the previous proposal distribution.  
Then the importance weight can be written as
\begin{equation*}
    \begin{split}
        \widetilde{W}_{t, 2} 
        = & \frac{p (U_{1:t}, \btheta \mid H_{t, 2} )}{q (U_{1:t}, \btheta \mid H_{t, 2} )} \\
        \propto & p ( U_{1:t}, \btheta \mid H_{t, 1} ) \cdot 
        p ( I_{t, 1} U_{t} \mid U_{t}, I_{t, 1} ) \cdot 
        \brck{ q (U_{1:t}, \btheta \mid H_{t, 1} ) }^{-1}.
    \end{split}
\end{equation*}
For any latent state $U_{t} = u_{t}$, we have  
$
p ( I_{t, 1} U_{t} = 0 \mid U_{t} = u_{t}, I_{t, 1} = 0 ) = 1.
$.
Then we have
\begin{equation*}
    \begin{split}
        \widetilde{W}_{t, 2} 
        \propto & \frac{p ( U_{1:t}, \btheta \mid H_{t, 1} )}{q (U_{1:t}, \btheta \mid H_{t, 1} )} \cdot
        p ( I_{t, 1} U_{t} \mid U_{t}, I_{t, 1} )
        = W_{t, 1},
    \end{split}
\end{equation*}
where 
\[
W_{t, 1} = \frac{p ( U_{1:t} \mid H_{t, 1} )}{q (U_{1:t} \mid H_{t, 1} )}
\]
is the weight for the marginal posterior distribution of $U_{1:t}$ given the history $H_{t, 1}$.  
Thus, the proposal distribution and the particle weights remain the same as at time $(t, 1)$.

When $I_{t} = 1$, we do not need the parameter $\btheta$ to approximate the belief state of the latent state.  
The posterior distribution $p (U_{1:t} \mid H_{t, 2} )$ can be expressed as
\begin{equation*}
    \begin{split}
        p (U_{1:t} \mid H_{t, 2} )
        = & p ( U_{1:t} \mid H_{t - 1, 2}, A_{t - 1, 2}, Z_{t}, O_{t}, I_{t, 1}, I_{t, 1} U_{t} ) \\
        = & p ( U_{1:t-1} \mid H_{t - 1, 2}, A_{t - 1, 2}, I_{t, 1} ) \cdot \\
        & p ( Z_{t} \mid U_{1:t-1}, H_{t - 1, 2}, A_{t - 1, 2}, I_{t, 1} ) \cdot \\
        & p ( I_{t, 1} U_{t} \mid Z_{t}, U_{1:t-1}, H_{t - 1, 2}, A_{t - 1, 2}, I_{t, 1} ) \cdot \\
        & p ( U_{t} \mid I_{t, 1} U_{t}, Z_{t}, U_{1:t-1}, H_{t - 1, 2}, A_{t - 1, 2}, I_{t, 1} ) \cdot \\
        & p ( O_{t} \mid U_{t}, Z_{t}, I_{t, 1} U_{t}, U_{1:t-1}, H_{t - 1, 2}, A_{t - 1, 2}, I_{t, 1} ) \cdot \\
        & \brck{ p ( Z_{t}, O_{t}, I_{t, 1} U_{t} \mid H_{t - 1, 2}, A_{t - 1, 2}, I_{t, 1} ) }^{-1} \\
        \propto & p ( U_{1:t-1} \mid H_{t - 1, 2} ) \cdot
        p ( Z_{t} \mid U_{t-1}, Z_{t - 1}, I_{t - 1, 1}, A_{t - 1, 2} ) \cdot \\
        & p ( I_{t, 1} U_{t} \mid Z_{t}, U_{t-1}, Z_{t - 1}, I_{t - 1, 1}, A_{t - 1, 2}, I_{t, 1} ) \cdot \\
        & p ( U_{t} \mid I_{t, 1} U_{t}, Z_{t}, U_{t-1}, Z_{t - 1}, I_{t - 1, 1}, A_{t - 1, 2}, I_{t, 1} ) \cdot 
        p ( O_{t} \mid U_{t} ).
    \end{split}
\end{equation*}
When the true latent state $U_{t} = u_{t}$, we have  
\begin{equation*}
    p ( I_{t, 1} U_{t} = u \mid U_{t} = u_{t}, I_{t, 1} = 1 ) =
    \delta (u - u_{t}),
\end{equation*}
according to the definition of the emission model.  
Therefore,  
\begin{equation*}
    \begin{split}
        & p ( U_{t} = u_{t} \mid I_{t, 1} U_{t} = u, Z_{t}, U_{t-1}, Z_{t - 1}, I_{t - 1, 1}, A_{t - 1, 2}, I_{t, 1} = 1 ) \\
        = & \frac{p ( I_{t, 1} U_{t} = u \mid U_{t} = u_{t}, I_{t, 1} = 1 ) \,
        p ( U_{t} = u_{t} \mid Z_{t}, U_{t-1}, Z_{t - 1}, I_{t - 1, 1}, A_{t - 1, 2}, I_{t, 1} = 1 )}
        {p ( I_{t, 1} U_{t} = u \mid Z_{t}, U_{t-1}, Z_{t - 1}, I_{t - 1, 1}, A_{t - 1, 2}, I_{t, 1} = 1 )} \\
        = & \delta (u - u_{t}).
    \end{split}
\end{equation*}
Suppose the proposal distribution can be factorized as
\begin{equation*}
    q (U_{1:t} \mid H_{t, 2} ) 
    = q (U_{1:t-1} \mid H_{t-1, 2}) \,
      q (U_t \mid U_{1:t-1}, H_{t, 2}).
\end{equation*}
Now if we take 
\begin{align*}
    q (U_t \mid U_{1:t-1}, H_{t, 2}) 
    = p ( U_{t} \mid I_{t, 1} U_{t}, Z_{t}, U_{t-1}, Z_{t - 1}, I_{t - 1, 1}, A_{t - 1, 2}, I_{t, 1} ), 
\end{align*}
the importance weight can be written as
\begin{equation*}
    \begin{split}
        \widetilde{W}_{t, 2} 
        \propto & \frac{p ( U_{1:t-1} \mid H_{t - 1, 2} )}{q ( U_{1:t-1} \mid H_{t - 1, 2} )} \cdot
        \frac{
        p ( U_{t} \mid I_{t, 1} U_{t}, Z_{t}, U_{t-1}, Z_{t - 1}, I_{t - 1, 1}, A_{t - 1, 2}, I_{t, 1} )
        }{
        q (U_t \mid U_{1:t-1}, H_{t, 2})
        } \cdot \\
        & p ( Z_{t} \mid U_{t-1}, Z_{t - 1}, I_{t - 1, 1}, A_{t - 1, 2} ) \cdot 
        p ( I_{t, 1} U_{t} \mid Z_{t}, U_{t-1}, Z_{t - 1}, I_{t - 1, 1}, A_{t - 1, 2}, I_{t, 1} ) \cdot
        p ( O_{t} \mid U_{t} ) \\
        \propto & W_{t-1, 2} \cdot
        p ( Z_{t} \mid U_{t-1}, Z_{t - 1}, I_{t - 1, 1}, A_{t - 1, 2} ) \cdot 
        p ( I_{t, 1} U_{t} \mid Z_{t}, U_{t-1}, Z_{t - 1}, I_{t - 1, 1}, A_{t - 1, 2}, I_{t, 1} ) \cdot
        p ( O_{t} \mid U_{t} ).
    \end{split}
\end{equation*}
In this case, the proposal distribution $q (U_t \mid U_{1:t-1}, H_{t, 2})$ simplifies to
\begin{equation*}
    q (U_t = u \mid U_{1:t-1}, H_{t, 1}, I_{t, 1} = 1, I_{t, 1} U_{t} = u_{t}) 
    = \delta (u - u_{t}),
\end{equation*}
which places all mass on $U_t = u_{t}$.  
This indicates that when $I_{t, 1} = 1$, we should always draw $\widehat{U}_{t, 2}^{(j)} = u_{t}$.  
Then, since $\widehat{U}_{t, 2}^{(j)}$ has the same value for all particles $j$, the likelihood term $p ( O_{t} \mid \widehat{U}_{t, 2}^{(j)} )$ is identical across all particles in the importance weight.  
Now we have
\begin{equation*}
    \begin{split}
        & p ( I_{t, 1} U_{t} = u_t \mid Z_{t}, U_{t-1}, Z_{t - 1}, I_{t - 1, 1}, A_{t - 1, 2}, I_{t, 1} ) \\
        = & \int p ( I_{t, 1} U_{t} = u' \mid U_{t} = u, Z_{t}, U_{t-1}, Z_{t - 1}, I_{t - 1, 1}, A_{t - 1, 2}, I_{t, 1} = 1 ) \,
        p ( U_{t} = u \mid Z_{t}, U_{t-1}, Z_{t - 1}, I_{t - 1, 1}, A_{t - 1, 2}, I_{t, 1} ) \, d u \\
        = & \int \delta(u_t - u) \,
        p ( U_{t} = u \mid Z_{t}, U_{t-1}, Z_{t - 1}, I_{t - 1, 1}, A_{t - 1, 2} ) \, d u \\
        = & p ( U_{t} = u_t \mid Z_{t}, U_{t-1}, Z_{t - 1}, I_{t - 1, 1}, A_{t - 1, 2} ).
    \end{split}
\end{equation*}
Therefore, after normalization, we obtain
\begin{equation*}
    \widetilde{W}_{t, 2}^{(j)} \propto 
    W_{t - 1, 2}^{(j)} \cdot 
    p ( Z_{t} = z_t, U_{t} = u_t \mid U_{t-1}, Z_{t - 1}, I_{t - 1, 1}, A_{t - 1, 2} ),
\end{equation*}
when we observe $Z_{t} = z_t$ and $I_{t, 1} U_{t} = u_t$.  

Note that the decomposition of $p (U_{1:t} \mid H_{t, 2} )$ from $H_{t - 1, 2}$ instead of $H_{t, 1}$ allows us to resample the particles of $U_t$.  
Otherwise, if the true value $u_t$ were not drawn from the proposal distribution $q(U_{1:t}, \btheta \mid H_{t, 1})$ at time $(t, 1)$, then all particle weights $\widetilde{W}_{t, 2}^{(j)}$ would be zero.

\section{Additional Algorithm Details} \label{sec:algorithm.details}

In this section, we provide implementation details of the active-measure algorithm.

\subsection{Active-Measure}

Algorithm~\ref{alg:active.measure} provides a full description of the Active-Measure procedure.

\begin{algorithm}[!htbp]
    \caption{Active-Measure}
    \label{alg:active.measure}
    \textbf{Input}: Hyperparameters $\lambda^I, (\sigma^I)^2, \lambda^A, (\sigma^A)^2, C$.
    
    \begin{algorithmic}[1] 
    \STATE Observe $Z_{1}$ and $O_{1}$. Construct the belief state $b^U_{1, 1}$ for $U_{1}$ using Algorithm~\ref{alg:belief.state.1}.
    \FOR{$t \ge 1$}
        \STATE Set $I_{t, 1} = 1$ if $t = 1$; otherwise set $I_{t, 1} = \argmax_{i \in \cI} \phi^I (b^{S^I}_{t, 1}, i)^{\top} \widetilde{\bbeta}^I_{t}$.  
        \STATE Observe $I_{t, 1} U_{t}$.  
        Update the belief state $b^U_{t, 2}$ for $U_{t}$ using Algorithm~\ref{alg:belief.state.2}.  
        \STATE Draw $\widetilde{\bbeta}^I_{t} \sim N (\bmu^I_{t}, \bSigma^I_{t})$ using~\eqref{equ:blr.posterior}.  
        \STATE Set $A_{t, 2} \sim \operatorname{Bernoulli}(0.5)$ if $t = 1$; otherwise set 
        $
        A_{t, 2} = \argmax_{a \in \cA} \phi^A (b^{S^A}_{t, 2}, a)^{\top} \widetilde{\bbeta}^A_{t}.
        $
        \STATE Observe $Z_{t + 1}$ and $O_{t + 1}$.  
        Construct the belief state $b^U_{t + 1, 1}$ for $U_{t + 1}$ using Algorithm~\ref{alg:belief.state.1}.  
        \STATE Draw $\widetilde{\bbeta}^A_{t} \sim N (\bmu^A_{t}, \bSigma^A_{t})$ using~\eqref{equ:blr.posterior}.  
    \ENDFOR
    \end{algorithmic}
\end{algorithm}

\subsection{RLSVI} \label{sec:rlsvi}

For completeness, we describe the standard RLSVI algorithm \citep{osband2016generalization} in Algorithm~\ref{alg:stationary.rlsvi}.  
Here we assume a stationary MDP setting, where the state is $S_t$, the action is $A_t$, and the reward is $R_t$.  

Define 
\begin{equation*}
\begin{split}
    \bX_{l} &= \phi (s_l, a_l), \\
    Y_l &= r_l + \gamma \max_a \phi (s_{l + 1}, a)^{\top} \widetilde{\bbeta}_{t - 1},
\end{split}
\end{equation*}
for $l \in \{1:t\}$,  
where $\phi$ is the basis function.  
Let $\Xb := [\bX_{1:t}]^{\top}$ and $\Yb := [Y_{1:t}]^{\top}$.  
We fit a BLR model for $\Yb$ given $\widetilde{\bbeta}_{t - 1}$.  
The posterior of $\bbeta_{t}$ is $N (\bmu_{t}, \bSigma_{t})$, where 
\begin{equation*} 
\begin{split}
        \bSigma_t &= \big[ (\Xb_t)^{\top} \Xb_t / \sigma^2 + \lambda \Ib \big]^{-1}, \\
        \bmu_t &= \bSigma_t \, [(\Xb_t)^{\top} \Yb_t / \sigma^2].
    \end{split}
\end{equation*}
An estimate of $\bbeta_{t}$ is then obtained by drawing $\widetilde{\bbeta}_{t} \sim N (\bmu_{t}, \bSigma_{t})$ from the posterior distribution.

\begin{algorithm}[!htbp]
    \caption{Stationary RLSVI}
    \label{alg:stationary.rlsvi}
    \textbf{Input}: Hyperparameters $\lambda, \sigma^2$, and initialization $\widetilde{\bbeta}_{0} = \bzero$.
    
    \begin{algorithmic}[1] 
    \STATE Observe the initial state $s_1$.
    \FOR{$t \ge 1$}
        \STATE Draw $\widetilde{\bbeta}_{t} \sim N (\bmu_{t}, \bSigma_{t})$ based on the previous parameter estimated $\widetilde{\bbeta}_{t - 1}$.
        \STATE Select the action $a_t = \argmax_{a} \phi (s_{t}, a)^{\top} \widetilde{\bbeta}_{t}$.
        \STATE Observe the reward $r_t$ and the next state $s_{t + 1}$.
    \ENDFOR
    \end{algorithmic}
\end{algorithm}

\section{Application Details} \label{sec:simulation.details}

In this section, we provide details for the HeartSteps application.

\subsection{HeartSteps as a Special Case of an AOMDP}  \label{sec:heartsteps.aomdp}

\paragraph{Definition.}
In HeartSteps, the reward $R_{t}$ corresponds to the next latent state $U_{t + 1}$, and the proximal outcome and engagement $[M_{t, 2}, E_{t}]$ form the observed state $Z_{t+1}$.  
The emission $O_t$ of the latent state $U_{t}$ is indexed as the emission $O_{t-1}$ of the latent reward $R_{t-1}$.  
The reward $r(Z_{t + 1}, U_{t + 1})$ of the AOMDP is defined as $r(M_{t, 2}, E_{t}, R_{t}) = R_t$.

\paragraph{State Construction.}
Lemma~\ref{lem:aomdp.periodic.pomdp} states that the state of $I_{t, 1}$ is $S^I_{t, 1} = [Z_{t}, U_{t}]$, which includes $[M_{t - 1, 2}, E_{t - 1}, R_{t - 1}]$.  
However, the causal DAG suggests that $M_{t - 1, 2}$ is independent of future rewards and states given $E_{t - 1}, R_{t - 1}$, and thus can be omitted from the state without affecting the optimal value function \citep{gao2025harnessing}.  
Notice that a context variable $C_{t, 2}$ is observed between $I_{t, 1}$ and $A_{t, 2}$, which does not exist in the general AOMDP framework.  
However, since $C_{t, 2}$ is exogenous and independent of other variables given $M_{t, 2}$, it does not affect the belief propagation and only becomes part of the state $S^A_{t, 2}$ of $A_{t, 2}$.  
Therefore, the optimal state for $I_{t}$ is $S^I_{t, 1} = [E_{t-1}, R_{t-1}]$, and the optimal state for $A_t$ is $S^A_{t, 2} = [E_{t-1}, R_{t-1}, C_{t, 2}, I_{t, 1}]$.

\paragraph{SMC.}
In Algorithms~\ref{alg:belief.state.1} and~\ref{alg:belief.state.2}, the probability 
$p ( z_{t} \mid \widehat{\btheta}^{(j)}, \widetilde{u}_{t-1, 2}^{(j)}, z_{t-1}, i_{t - 1, 1}, a_{t - 1, 2} )$ 
is used to update the particle weight.  
However, the causal DAG implies that $E_{t - 1} \indep R_{t - 2} \mid E_{t - 2}, I_{t - 1, 1}, A_{t - 1, 2}$.  
Therefore, the belief states $b^{U}_{t, 1}$ and $b^{U}_{t, 2}$ do not depend on $E_{t - 1}$.  
Only $M_{t - 1, 2}$ is needed to update the particle weight (see details in Appendix~\ref{sec:posterior.parameters}).

\paragraph{Basis Functions and Reward Functions.}
When a belief state is approximately normal, the mean and standard deviation of the particles serve as sufficient statistics of the normal distribution.  
Let $\{\widehat{r}_{t, k}^{(j)}\}_{j=1}^J$ be the particles and $\widehat{b}^R_{t, k} (u) = \sum_{j=1}^J w_{t, k}^{(j)} \delta (r - \widehat{r}_{t, k}^{(j)})$ be the estimated belief state of the latent reward $R_{t}$ at time $(t, k)$ for $k = 1, 2$.
We thus define 
\begin{align*}
    \bar{b}^{R}_{t, k} := & \, \bbE_{\hat{b}^{R}_{t, k}} (s) = \sum_{j=1}^J w_{t, k}^{(j)} \widehat{r}_{t, k}^{(j)}, \\
    \widetilde{b}^{R}_{t, k} := & \, \operatorname{Std}_{\hat{b}^{R}_{t, k}} (s) = \brce{\sum_{j=1}^J w_{t, k}^{(j)} \big(\widehat{r}_{t, k}^{(j)} - \bar{b}^{R}_{t, k}\big)^2}^{1/2}
\end{align*}
as the expectation and standard deviation under the belief state.  
The basis functions are constructed as  
\begin{align*}
    \phi^I (b^{S^I}_{t, 1}, I_{t, 1}) = & [1, E_{t - 1}, \bar{b}^{R}_{t - 1, 1}, \widetilde{b}^{R}_{t - 1, 1}, I_{t, 1}, I_{t, 1} E_{t - 1}, I_{t - 1, 1} \bar{b}^{R}_{t - 1, 1}], \\
    \phi^A (b^{S^A}_{t, 2}, A_{t, 2}) = & [1, E_{t - 1}, \bar{b}^{R}_{t - 1, 2}, C_{t, 2}, I_{t, 1}, A_{t, 2}, A_{t, 2} E_{t - 1}, A_{t, 2} \bar{b}^{R}_{t - 1, 2}, A_{t, 2} C_{t, 2}].
\end{align*}
The standard deviation $\widetilde{b}^{R}_{t - 1, 2}$ is not included in $\phi^A$ since it is highly correlated with $I_{t, 1}$.  

Remember that the reward is defined as $r(Z_{t + 1}, U_{t + 1, 1}) = R_t$, and the target for the control action in RLSVI is  $r(Z_{t + 1}, b^{U}_{t + 1, 1}) = \int r(Z_{t + 1}, u) b^{U}_{t + 1, 1} (u) du$.
Then it can be estimated as $r(Z_{t + 1}, \widehat{b}^{U}_{t + 1, 1}) = \sum_{j=1}^J w_{t, 1}^{(j)} \widehat{r}_{t, 2}^{(j)} = \bar{b}^{R}_{t, 2}$.

\subsection{Update the Posterior of Unknown Parameters} \label{sec:posterior.parameters}

Particle learning \citep{storvik2002particle,carvalho2010particle} can be inefficient in a general POMDP where the posterior distribution of the parameters $\btheta$ is intractable.  
However, with certain working models, the posterior distribution of $\btheta$ admits a closed form.  
For example, models in the exponential family with conjugate priors yield closed-form posteriors, including linear models with Gaussian noise, binomial models, multinomial models, and Poisson models.

Here, we use a working model with linear mean and Gaussian noise.  
Specifically, suppose the mean of each variable is a linear function of its parents in the causal DAG, and the noise follows a Gaussian distribution.  
That is,
\begin{equation} \label{equ:testbed.model}
\begin{split}
    M_{t, 2} &= \theta^{M}_{0} + \theta^{M}_{1} E_{t-1} + \theta^{M}_{2} R_{t-1} + \theta^{M}_{3} C_{t, 2} 
    + A_{t, 2} (\theta^{M}_{4} + \theta^{M}_{5} E_{t-1} + \theta^{M}_{6} R_{t-1} + \theta^{M}_{7} C_{t, 2}) + \epsilon^M_{t, 2}, \\
    R_{t} &= \theta^{R}_{0} + \theta^{R}_{1} M_{t, 2} + \theta^{R}_{2} E_{t} + \theta^{R}_{3} R_{t-1} + \epsilon^R_{t}, \\
    O_{t} &= \theta^{O}_{0} + \theta^{O}_{1} R_{t} + \epsilon^O_{t},
\end{split}
\end{equation}
where 
$\btheta^M = \theta^{M}_{0:7}$, 
$\btheta^R = \theta^{R}_{0:3}$, and
$\btheta^O = \theta^{O}_{0:1}$.  
Let $\btheta := \{ \btheta^M, \btheta^R, \btheta^O \}$ denote all the parameters used in SMC.  
The noise terms $\epsilon^M_{t, 2}$, $\epsilon^R_{t}$, and $\epsilon^O_{t}$ follow Gaussian distributions with mean zero and fixed variances $\sigma^{2M}$, $\sigma^{2R}$, and $\sigma^{2O}$, respectively.  
Suppose the prior of $\btheta$ is $\btheta^V \sim N (\bnu_0^V, \bLambda_0^V)$ for $V \in \{M, R, O\}$.  

Given the history 
\[
h_{t,2} = [E_0, O_0, I_{1, 1}, I_{1, 1} R_0, C_{1, 2}, A_{1, 2}, M_{1, 2}, E_1, O_1, \dots, E_{t - 1}, O_{t - 1}, I_{t, 1}, I_{t, 1} R_{t - 1}],
\]
and a particle value $\widehat{u}_{1:t, 2}^{(j)} = \widehat{r}_{0:t-1, 2}^{(j)}$, define
\begin{align*}
    \Xb^M_{t} := & [X^M_{1:t-1}]^{\top}, 
    && \text{with } X^M_{l} := [1, E_{l - 1}, \widehat{r}_{l - 1, 2}^{(j)}, C_{l, 2}, A_{l, 2}, A_{l, 2} E_{l - 1}, A_{l, 2} \widehat{r}_{l - 1, 2}^{(j)}, A_{l, 2} C_{l, 2}], \text{ for } l = 1, \dots, t - 1, \\
    \Yb^M_{t} := & [M_{1:t-1, 2}]^{\top}, \\
    \Xb^R_{t} := & [X^R_{1:t-1}]^{\top}, 
    && \text{with } X^R_{l} := [1, M_{l, 2}, E_t, \widehat{r}_{l - 1, 2}^{(j)}], \text{ for } l = 1, \dots, t - 1, \\
    \Yb^R_{t} := & [\widehat{r}_{1:t-1, 2}^{(j)}]^{\top}, \\
    \Xb^O_{t} := & [X^O_{1:t-1}]^{\top}, 
    && \text{with } X^O_{l} := [1, \widehat{r}_{l, 2}^{(j)}], \text{ for } l = 1, \dots, t - 1, \\
    \Yb^O_{t} := & [O_{1:t-1}]^{\top}.
\end{align*}

Under the working model \eqref{equ:testbed.model}, the posterior distribution of $\btheta^V$ is 
$
\btheta^V \mid \widehat{u}_{1:t, 2}^{(j)}, h_{t,2} \sim N (\bnu_t^V, \bLambda_t^V),
$
for $V \in \{M, R, O\}$, where
\begin{align*}
    \bLambda_t^M &= \brce{\frac{1}{\sigma^{2}_R} (\Xb^M_{t})^{\top} \Xb^M_{t} + (\bLambda_0^M)^{-1}}^{-1}, \\
    \bnu_t^M &= \bLambda_t^M \brce{\frac{1}{\sigma^{2}_R} (\Xb^M_{t})^{\top} \Yb^M_{t} + (\bLambda_0^M)^{-1} \bnu_0^M }.
\end{align*}
Then, a particle 
$
\widehat{\btheta}_t^{(j)} = [\widehat{\btheta}_t^{M(j)}, \widehat{\btheta}_t^{R(j)}, \widehat{\btheta}_t^{O(j)}],
$
is drawn with 
\[
\widehat{\btheta}_t^{V(j)} \mid \widehat{u}_{1:t-1, 2}^{(j)}, h_{t-1,2} \sim N (\bnu_{t - 1}^V, \bLambda_{t - 1}^V),
\]
for $V \in \{M, R, O\}$.

As discussed in Appendix~\ref{sec:heartsteps.aomdp}, $E_{t - 1}$ is not needed when updating the belief particles.  
Given $\widehat{\btheta}_t^{(j)}$, we draw 
\[
\widetilde{r}_{t - 1, 1}^{(j)} \sim p ( r_{t - 1} \mid \widehat{\btheta}_t^{R(j)}, m_{t - 1, 2}, e_{t-1}, \widehat{r}_{t-2, 2}^{(j)} ),
\]
and update the particle weight as
\begin{equation*}
    \widetilde{w}_{t, 1}^{(j)} \propto 
    w_{t-1, 2}^{(j)} \,
    p ( m_{t - 1, 2} \mid \widehat{\btheta}_t^{M(j)}, \widetilde{r}_{t-2, 2}^{(j)}, e_{t - 2}, c_{t - 1, 2}, a_{t - 1, 2} ) \,
    p ( o_{t - 1} \mid \widehat{\btheta}_t^{O(j)}, \widetilde{r}_{t - 1, 1}^{(j)} ).
\end{equation*}

\subsection{Reward Design} \label{sec:reward.design}

When the effects of actions on rewards are mediated by the proximal outcome $M_{t, 2}$ and the engagement $E_t$, the causal effects become harder to detect.  
Fortunately, the mediators can be leveraged to construct improved rewards in the RL algorithm by following the idea of the surrogate index \citep{athey2019surrogate,yang2024targeting}.  
The conditional mean of the reward given the mediators has smaller variance than the original reward.  
Based on the working model \eqref{equ:testbed.model}, the mean of the reward $R_t$ can be estimated as  
$
\widehat{R}_t := [1, M_{t, 2}, E_t, \bar{b}_{t - 1, 2}^{R}] \, \widehat{\btheta}_t^{R(j)}.
$ 
We then use $\widehat{R}_t$ as the reward in Algorithm~\ref{alg:active.measure}.

\subsection{Hyperparameters} \label{sec:hyperparameters}

The prior mean $\bnu_0^V$ for $V \in \{M, R, O\}$ is estimated by pooling data across all users in HeartSteps V3, following the same procedure as in \citet{gao2025harnessing}.  
We obtain $\bnu_0^M = [-0.043, -0.026, 0.062, 0.418, 0.001, 0.003, -0.035, 0.011]$, $\bnu_0^R = [-0.005, 0.029, 0.012, 0.861]$, and $\bnu_0^O = [0.034, 0.534]$.  
Here, the second coordinate of $\bnu_0^R$ is computed by summing over the five original estimated parameters $\theta^{R}_{k}$ for the 5 interventions in HeartSteps V3.  
The prior covariance $\bLambda_0^V$ is set to a diagonal matrix $0.01 \Ib$, where $\Ib$ is the identity matrix.  
The noise variances $\sigma^{2M}$, $\sigma^{2R}$, and $\sigma^{2O}$ are also estimated from HeartSteps V3 as the variances of the residuals obtained by fitting linear regressions for $M$, $R$, and $O$, respectively. 
We have $\sigma^{2M} = 0.972$, $\sigma^{2R} = 0.240$, and $\sigma^{2O} = 0.637$.  
The same priors and noise variances are used for all users in our simulation experiments.  

The discount factor is chosen as $\gamma = 0.9$ to balance discount regularization and the modeling of long-term effects.  
In HeartSteps, the optimal action not only leads to a high instantaneous reward but also places the user in a promising state that yields higher rewards in the future.  
In behavioral science, this process is referred to as habit formation.  

The number of particles $J$ is set to 50, which is sufficient to approximate the belief state under the working model \eqref{equ:testbed.model}.  
The numerical experiments in \citet{lim2023optimality} also demonstrate that between $10^1$ and $10^2$ particles already yield good performance in simpler problems.  
The parameters of the target $\widetilde{\bbeta}^A_{t^-}$ are copied from $\widetilde{\bbeta}^A_{t}$ every $C = 10$ steps.  

When selecting the hyperparameters $\lambda^I, (\sigma^I)^2, \lambda^A, (\sigma^A)^2$, note that $\lambda \cdot \sigma^2$ is equivalent to the tuning parameter of an $L_2$ penalty.  
Therefore, we select $\lambda^I \cdot (\sigma^I)^2$ from $\{0.2, 0.5\}$, $(\sigma^I)^2$ from $\{0.02, 0.1\}$, $\lambda^A \cdot (\sigma^A)^2$ from $\{5, 20\}$, and $(\sigma^A)^2$ from $\{0.02, 0.1\}$.

\subsection{Simulation Testbed} \label{sec:testbed}

We construct our simulation testbed based on the public testbed developed by \citet{gao2025harnessing}.  
The original testbed includes five decision times per day.  
To focus on the discussion of active measuring, we consider a simplified setting with only one control action $A_{t, 2}$ per day.  
Adapting the original testbed to the problem described in Figure~\ref{fig:dag}, we aggregate the effects of the five actions $A_{t, 1:5}$ on both the reward and engagement, effectively treating all five actions as identical.  
Furthermore, we introduce an additional effect of $I_{t, 1}$ on the engagement $E_t$.  
Specifically, we modify equation (38) in \citet{gao2025harnessing} as follows:
\begin{equation*}
\begin{split}
    C_{t, 2} &= \theta^C_{0} + \epsilon^C_{t, 2}, \\
    M_{t, 2} &= \theta^{M}_{0} + \theta^{M}_{1} E_{t-1} + \theta^{M}_{2} R_{t-1} + \theta^{M}_{3} C_{t, 2} 
    + A_{t, 2} (\theta^{M}_{4} + \theta^{M}_{5} E_{t-1} + \theta^{M}_{6} R_{t-1} + \theta^{M}_{7} C_{t, 2}) + \epsilon^M_{t, 2}, \\
    E_{t} &= \theta^{E}_{0} + \theta^{E}_{1} E_{t-1} + \Big(\sum_{k=1}^5 \theta^{E}_{k+1}\Big) A_{t,2} 
    + \Big(\sum_{k=1}^5 \theta^{E}_{k + 6}\Big) A_{t,2} E_{t-1} + \theta^{E}_{I} I_{t,1} + \theta^{E}_{IE} I_{t,1} E_{t-1} + \epsilon^E_{t}, \\
    R_{t} &= \theta^{R}_{0} + \Big(\sum_{k=1}^5 \theta^{R}_{k}\Big) M_{t,2} + \theta^{R}_{6} E_{t} + \theta^{R}_{7} R_{t-1} + \epsilon^R_{t}, \\
    O_{t} &= \theta^{O}_{0} + \theta^{O}_{1} R_{t} + \epsilon^O_{t},
\end{split}
\end{equation*}
where $\theta^{E}_{I}$ and $\theta^{E}_{IE}$ are manually set, while all other parameters are taken from the HeartSteps testbed.  
The noise terms $\epsilon^C_{t, 2}$, $\epsilon^M_{t, 2}$, $\epsilon^E_{t}$, $\epsilon^R_{t}$, and $\epsilon^O_{t}$ have mean zero and variances $\sigma^{2C}$, $\sigma^{2M}$, $\sigma^{2E}$, $\sigma^{2R}$, and $\sigma^{2O}$, respectively.  
In addition, to ensure that engagement has a positive effect on the reward, we clip $\theta^{R}_{6}$ as $\max\{\theta^{R}_{6}, 0.02\}$.  

The effect size of the positive effect from $A_{t, 2} R_{t - 1}$ to $R_t$ through $M_{t, 2}$ is 
$
\theta^{M}_{6} (\sum_{k=1}^5 \theta^{R}_{k}) / \sqrt{\sigma^{2R} + \sigma^{2M}(\sum_{k=1}^5 \theta^{R}_{k})^2}
$.
The vanilla testbed, constructed directly from the HeartSteps dataset, has the minimal effect size.  
To examine the performance of the proposed algorithm across different testbed variants, we modify the parameter $\theta^{M}_{6}$ to 0.5 or 0.8 to achieve small and medium effect sizes.  
The average effect sizes across all users for the minimal, small, and medium positive effects are 0.026, 0.119, and 0.191, respectively.  

The effect size of the negative effect from $I_{t, 1}$ to $R_t$ through $E_{t}$ is 
$
\theta^{E}_{I} \theta^{R}_{6} / \sqrt{\sigma^{2R} + \sigma^{2E}(\theta^{R}_{6})^2}
$.
Since the measure action was not taken in HeartSteps V2, the vanilla testbed does not contain this negative effect.  
To create testbed variants with minimal and small effect sizes, we adjust the parameters $\theta^{E}_{I}$ and $\theta^{E}_{IE}$.  
The average effect sizes across all users for the zero, minimal, and small negative effects are 0, 0.010, and 0.039, respectively.

\subsection{Details of Always-Measure and Never-Measure} \label{sec:always.never.measure}

Always-measure and never-measure algorithms set $I_{t, 1}$ to one or zero with probability one and choose $A_{t, 2}$ using RLSVI.  
Specifically, the state is $S^A_{t, 2} = [E_{t-1}, R_{t-1}, C_{t, 2}]$, since $I_{t, 1}$ is a constant.  
Define the basis function 
\[
\phi^A (b^{S^A}_{t, 2}, A_{t, 2}) = 
[1, E_{t - 1}, \bar{b}^{U}_{t - 1, 2}, C_{t, 2}, A_{t, 2}, A_{t, 2} E_{t - 1}, A_{t, 2} \bar{b}^{U}_{t - 1, 2}, A_{t, 2} C_{t, 2}].
\]
Similar to the active-measure algorithm, define 
$\bX^A_{l} = \phi^A (b^{S^A}_{l, 2}, A_{l, 2})$ and
\begin{equation*}
    Y^A_{l} = r(Z_{l + 1}, b^{U}_{l + 1, 1}) + \gamma \phi^I (b^{S^A}_{l + 1, 2}, a')^{\top} \widetilde{\bbeta}^A_{t - 1}, 
    \quad \text{where} \quad
    a' = \argmax_{i \in \cI} \phi^I (b^{S^A}_{l + 1, 2}, i')^{\top} \widetilde{\bbeta}^A_{t^-}.
\end{equation*}
Here, $\widetilde{\bbeta}^A_{t^-}$ is copied from $\widetilde{\bbeta}^A_{t}$ every $C$ steps.  
We fit a BLR on $\Yb^A = [Y^A_{1:t}]^{\top}$ using $\Xb^A = [\bX^A_{1:t}]^{\top}$ to obtain the posterior distribution $N (\bmu^A_{t}, \bSigma^A_{t})$ of $\bbeta^A_{t}$,  
where 
\begin{equation} \label{equ:blr.posterior.always.never}
    \begin{split}
        \bSigma^A_t &= \big[ (\Xb^A_t)^{\top} \Xb^A_t / (\sigma^A)^2 + \lambda^A \Ib \big]^{-1}, \\
        \bmu^A_t &= \bSigma^A_t \, [(\Xb^A_t)^{\top} \Yb^A_t / (\sigma^A)^2].
    \end{split}
\end{equation}
An estimate of $\bbeta^A_{t}$ is then obtained by drawing $\widetilde{\bbeta}^A_{t} \sim N (\bmu^A_{t}, \bSigma^A_{t})$ from the posterior distribution.  

See Algorithm~\ref{alg:always.never.measure} for a full description of the always-measure and never-measure algorithms.  
The always-measure algorithm takes $P_0 = 1$, while the never-measure algorithm takes $P_0 = 0$.  
All other hyperparameters---$\lambda^A$, $(\sigma^A)^2$, $C$, $J$, and $\gamma$---and priors are set as described in Appendix~\ref{sec:hyperparameters}.

\begin{algorithm}[!htbp]
    \caption{Always-Measure or Never-Measure}
    \label{alg:always.never.measure}
    \textbf{Input}: Hyperparameters $\lambda^A, (\sigma^A)^2, C, P_0$.
    
    \begin{algorithmic}[1] 
    \STATE Observe $Z_{1}$ and $O_{1}$. Construct the belief state $b^U_{1, 1}$ for $U_{1}$ using Algorithm~\ref{alg:belief.state.1}.
    \FOR{$t \ge 1$}
        \STATE Set $I_{t, 1} = 1$ with probability $P_0$.          
        \STATE Observe $I_{t, 1} U_{t}$. Update the belief state $b^U_{t, 2}$ for $U_{t}$ using Algorithm~\ref{alg:belief.state.2}.  
        \STATE Take $A_{t, 2} \sim \operatorname{Bernoulli}(0.5)$ if $t = 1$; otherwise set 
        $
        A_{t, 2} = \argmax_{a \in \cA} \phi^A (b^{S^A}_{t, 2}, a)^{\top} \widetilde{\bbeta}^A_{t - 1}.
        $
        \STATE Observe $Z_{t + 1}$ and $O_{t + 1}$.  
        \STATE Draw $\widetilde{\bbeta}^A_{t} \sim N (\bmu^A_{t}, \bSigma^A_{t})$ using \eqref{equ:blr.posterior.always.never}.  
    \ENDFOR
    \end{algorithmic}
\end{algorithm}

\subsection{Details of Dyna-ATMQ} \label{sec:dyna.atmq}

In implementing the Dyna-ATMQ algorithm, we adapted the open-source BAM-QMDP implementation by \citet{krale2023act} to accommodate our testbed structure.  
We retained the core ATM loop, which selects a control action from a belief-weighted Q-table and separately evaluates whether to measure based on the predicted measuring value and cost.  
The Dirichlet-distribution-based transition function, the Dyna-framework, and the decoupled action–measurement decision rule were preserved from the original BAM-QMDP code, while several components were modified to fit our setting.  

We defined a 64-state discrete representation of the environment by discretizing $C_{t, 2}$, $E_{t - 1}$, and observed $R_{t - 1}$.  
Each variable was discretized into four bins corresponding to $x < -\sigma$, $-\sigma \le x < 0$, $0 \le x < \sigma$, and $x \ge \sigma$, where $x$ is the value of a continuous variable and $\sigma$ is its standard deviation.  
Since all variables in the testbed were standardized, the cutoff values were $-1$, $0$, and $1$.  

In addition, since Dyna-ATMQ only picks up the signal from an observed fixed cost, we need to treat the cost as a tuning parameter.
We select the cost from values of 0.0, 0.01, 0.02, and 0.05, and report the performance of the one with the highest cumulative reward.

For other hyperparameters, the number of particles is set to 100, the number of offline training steps is 5, and the discount factor is $\gamma = 0.9$.  
Based on our simulation results, the average cumulative reward of Dyna-ATMQ can be improved by adding an exploration phase at the beginning.  
Therefore, we include a warm-up period of 20 decision times, during which Dyna-ATMQ takes the control and measurement actions with probability 0.5.  

\subsection{Additional Simulation Results} \label{sec:additional.simulation.results}

\begin{figure*}
    \centering
    \begin{subfigure}{0.23\textwidth}
        \centering
        \includegraphics[width=\textwidth]{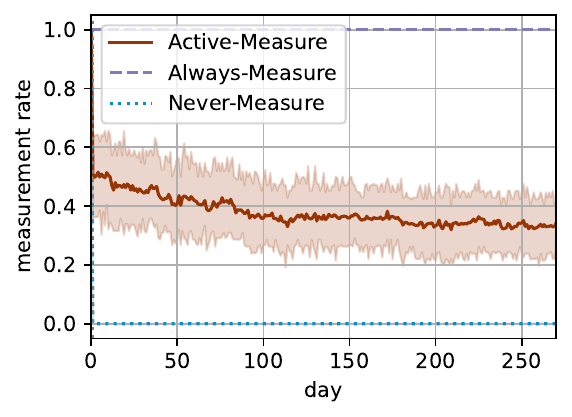}
        \caption{Minimal +, zero -}
        \label{subfig:s1.i}
    \end{subfigure}
    \hfill
    \begin{subfigure}{0.23\textwidth}
        \centering
        \includegraphics[width=\textwidth]{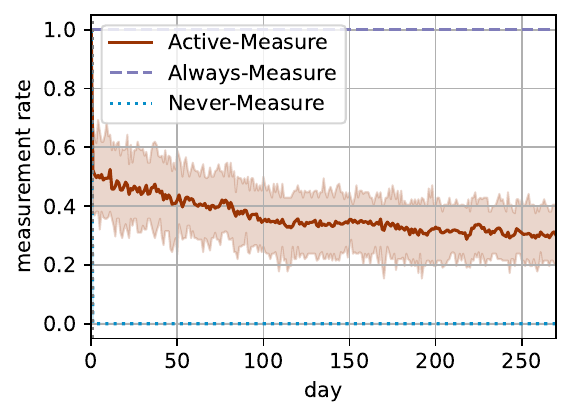}
        \caption{Minimal +, minimal -}
        \label{subfig:s4.i}
    \end{subfigure}
    \hfill
    \begin{subfigure}{0.23\textwidth}
        \centering
        \includegraphics[width=\textwidth]{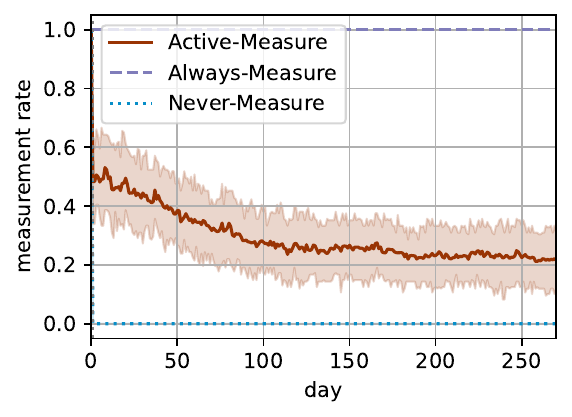}
        \caption{Minimal +, small -}
        \label{subfig:s7.i}
    \end{subfigure}
    \hfill
    \begin{subfigure}{0.23\textwidth}
        \centering
        \includegraphics[width=\textwidth]{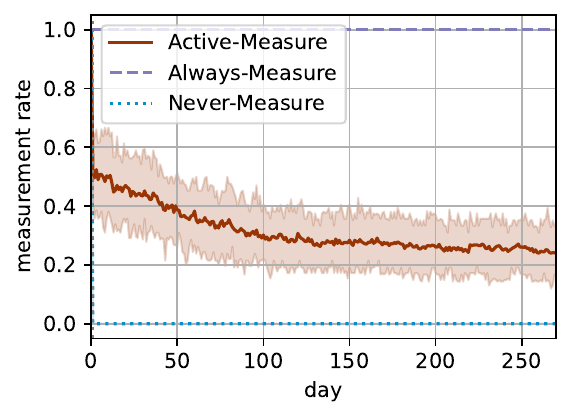}
        \caption{Minimal +, zero $O_t$}
        \label{subfig:s10.i}
    \end{subfigure}
    \hfill
    \begin{subfigure}{0.23\textwidth}
        \centering
        \includegraphics[width=\textwidth]{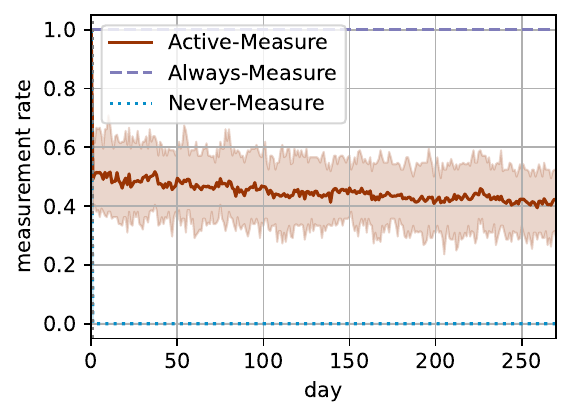}
        \caption{Small +, zero -}
        \label{subfig:s2.i}
    \end{subfigure}
    \hfill
    \begin{subfigure}{0.23\textwidth}
        \centering
        \includegraphics[width=\textwidth]{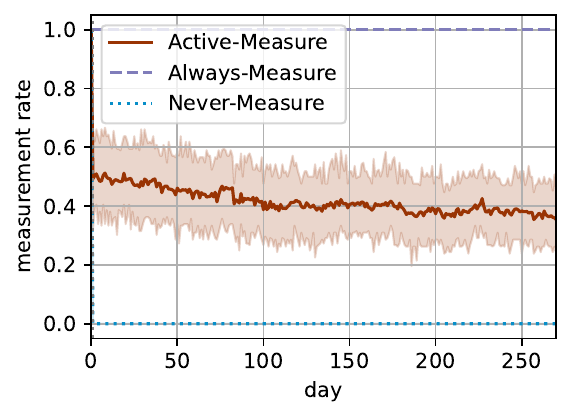}
        \caption{Small +, minimal -}
        \label{subfig:s5.i}
    \end{subfigure}
    \hfill
    \begin{subfigure}{0.23\textwidth}
        \centering
        \includegraphics[width=\textwidth]{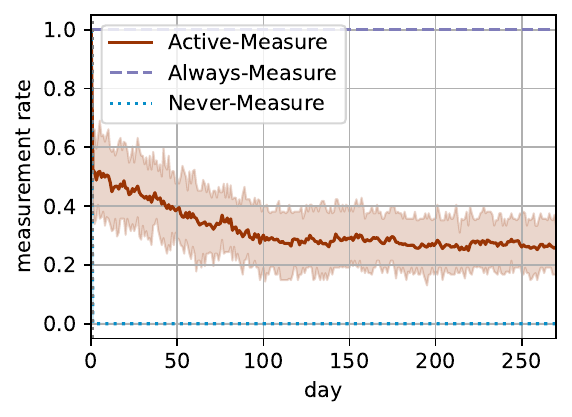}
        \caption{Small +, small -}
        \label{subfig:s8.i}
    \end{subfigure}
    \hfill
    \begin{subfigure}{0.23\textwidth}
        \centering
        \includegraphics[width=\textwidth]{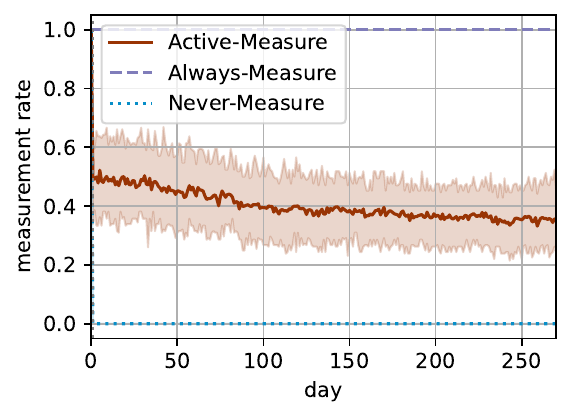}
        \caption{Small +, zero $O_t$}
        \label{subfig:s11.i}
    \end{subfigure}
    \hfill
    \begin{subfigure}{0.23\textwidth}
        \centering
        \includegraphics[width=\textwidth]{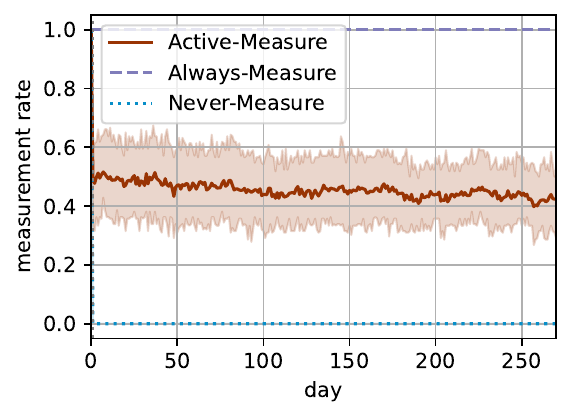}
        \caption{Medium +, zero -}
        \label{subfig:s3.i}
    \end{subfigure}
    \hfill
    \begin{subfigure}{0.23\textwidth}
        \centering
        \includegraphics[width=\textwidth]{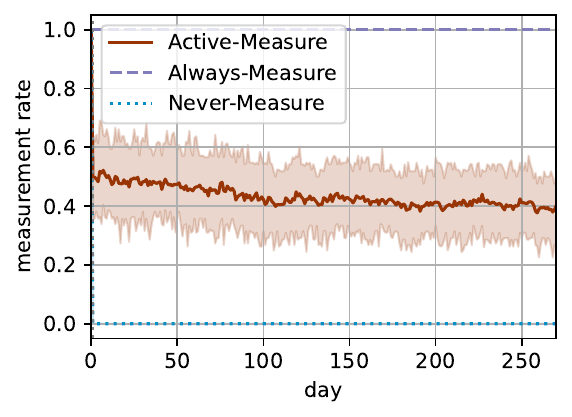}
        \caption{Medium +, minimal -}
        \label{subfig:s6.i}
    \end{subfigure}
    \hfill
    \begin{subfigure}{0.23\textwidth}
        \centering
        \includegraphics[width=\textwidth]{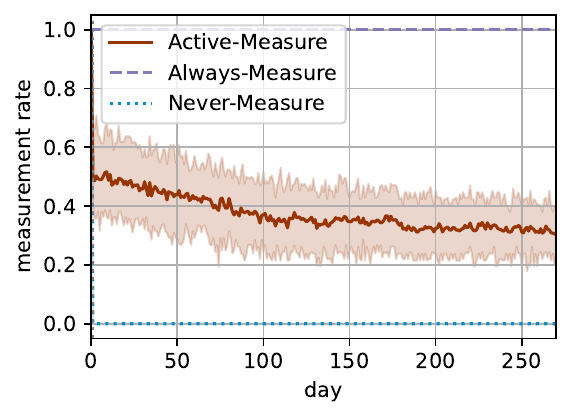}
        \caption{Medium +, small -}
        \label{subfig:s9.i}
    \end{subfigure}
    \hfill
    \begin{subfigure}{0.23\textwidth}
        \centering
        \includegraphics[width=\textwidth]{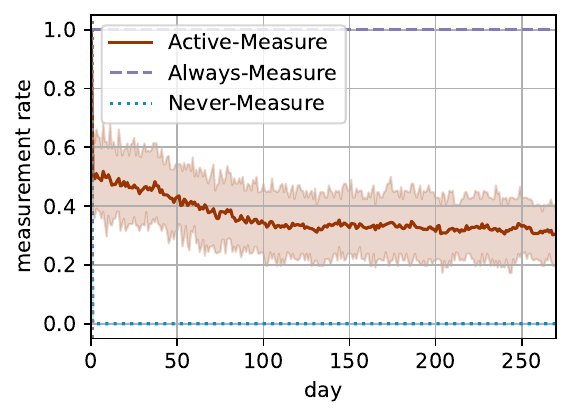}
        \caption{Medium +, zero $O_t$}
        \label{subfig:s12.i}
    \end{subfigure}
    \caption{The measurement rate.}
    \label{fig:simulation.results.i}
\end{figure*}

We report the measurement rate 
$
\frac{1}{42} \sum_{i = 1}^{42} \bbone (I_{t, i} = 1),
$
averaged across all users at each time $t$.  
Figure~\ref{fig:simulation.results.i} shows its mean and 95\% confidence interval over 50 replications.  
The experimental scenarios are the same as those in Figure~\ref{fig:simulation.results.r}.  
We observe that the measurement rate decreases as the negative effect increases.  
Moreover, under the same negative effect, the measurement rate increases as the positive effect increases, indicating that the benefits of measurement become greater.  

\begin{figure*}
    \centering
    \begin{subfigure}{0.23\textwidth}
        \centering
        \includegraphics[width=\textwidth]{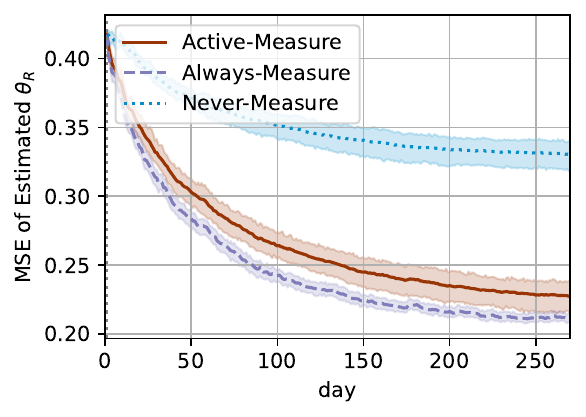}
        \caption{Minimal +, zero -}
        \label{subfig:s1.mse}
    \end{subfigure}
    \hfill
    \begin{subfigure}{0.23\textwidth}
        \centering
        \includegraphics[width=\textwidth]{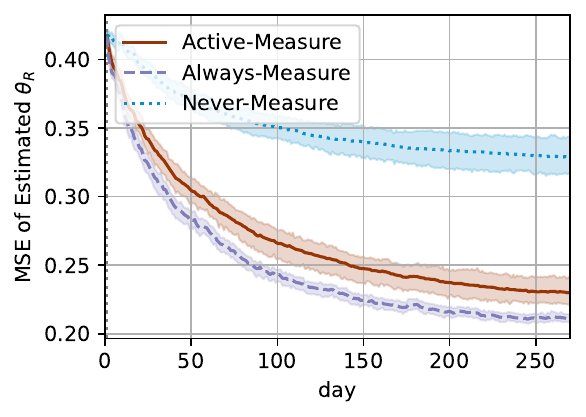}
        \caption{Minimal +, minimal -}
        \label{subfig:s4.mse}
    \end{subfigure}
    \hfill
    \begin{subfigure}{0.23\textwidth}
        \centering
        \includegraphics[width=\textwidth]{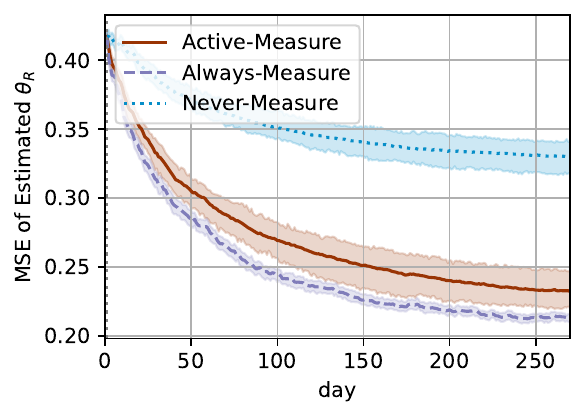}
        \caption{Minimal +, small -}
        \label{subfig:s7.mse}
    \end{subfigure}
    \hfill
    \begin{subfigure}{0.23\textwidth}
        \centering
        \includegraphics[width=\textwidth]{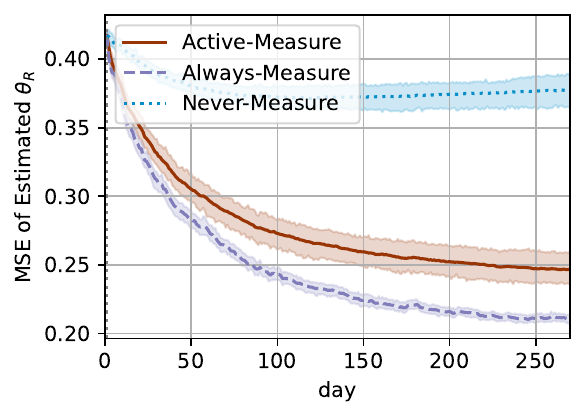}
        \caption{Minimal +, zero $O_t$}
        \label{subfig:s10.mse}
    \end{subfigure}
    \hfill
    \begin{subfigure}{0.23\textwidth}
        \centering
        \includegraphics[width=\textwidth]{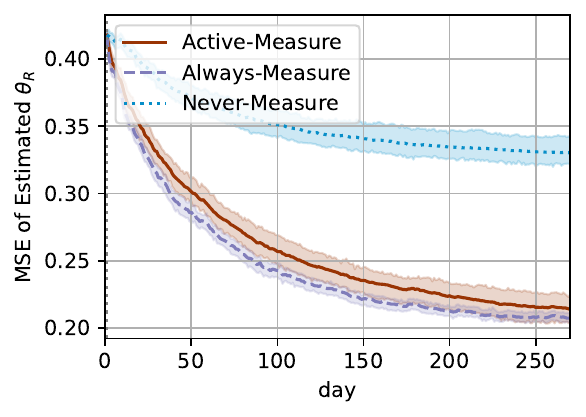}
        \caption{Small +, zero -}
        \label{subfig:s2.mse}
    \end{subfigure}
    \hfill
    \begin{subfigure}{0.23\textwidth}
        \centering
        \includegraphics[width=\textwidth]{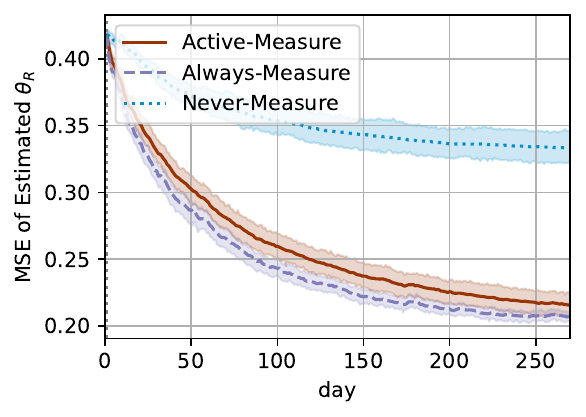}
        \caption{Small +, minimal -}
        \label{subfig:s5.mse}
    \end{subfigure}
    \hfill
    \begin{subfigure}{0.23\textwidth}
        \centering
        \includegraphics[width=\textwidth]{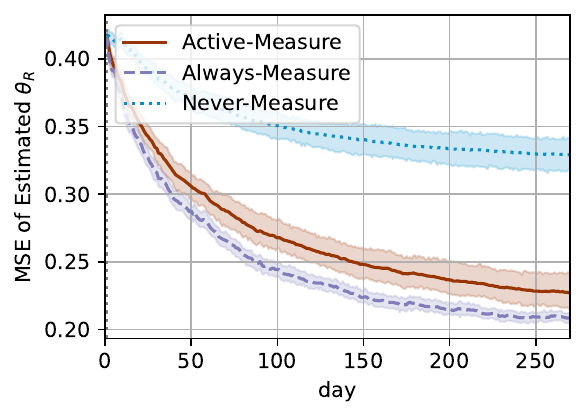}
        \caption{Small +, small -}
        \label{subfig:s8.mse}
    \end{subfigure}
    \hfill
    \begin{subfigure}{0.23\textwidth}
        \centering
        \includegraphics[width=\textwidth]{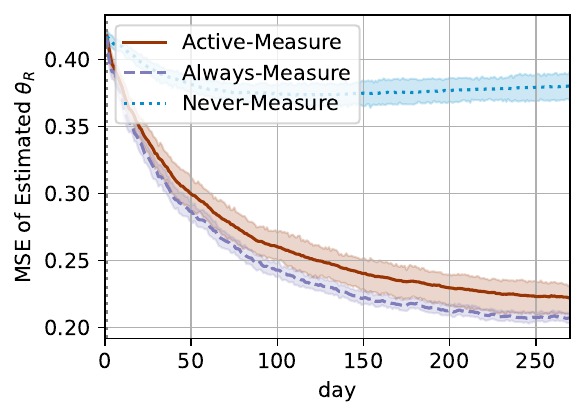}
        \caption{Small +, zero $O_t$}
        \label{subfig:s11.mse}
    \end{subfigure}
    \hfill
    \begin{subfigure}{0.23\textwidth}
        \centering
        \includegraphics[width=\textwidth]{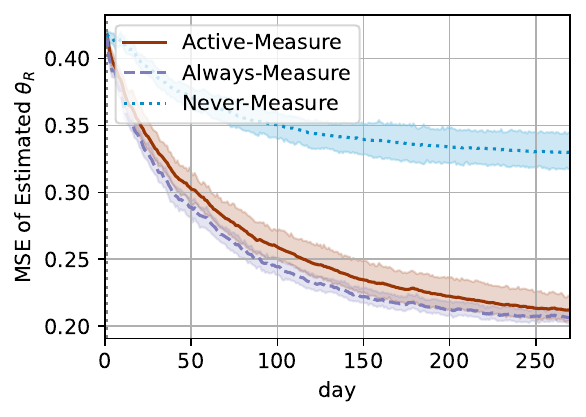}
        \caption{Medium +, zero -}
        \label{subfig:s3.mse}
    \end{subfigure}
    \hfill
    \begin{subfigure}{0.23\textwidth}
        \centering
        \includegraphics[width=\textwidth]{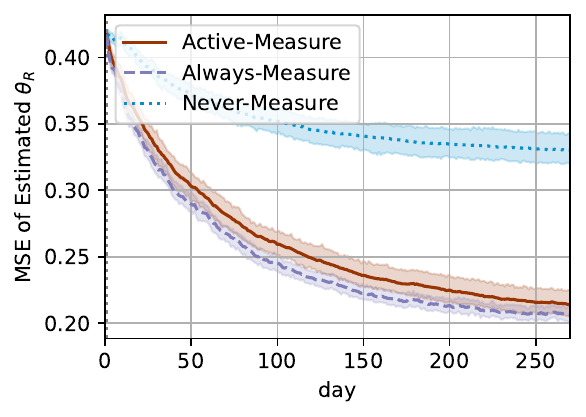}
        \caption{Medium +, minimal -}
        \label{subfig:s6.mse}
    \end{subfigure}
    \hfill
    \begin{subfigure}{0.23\textwidth}
        \centering
        \includegraphics[width=\textwidth]{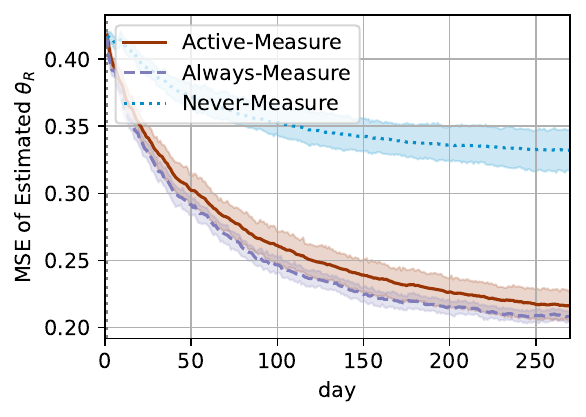}
        \caption{Medium +, small -}
        \label{subfig:s9.mse}
    \end{subfigure}
    \hfill
    \begin{subfigure}{0.23\textwidth}
        \centering
        \includegraphics[width=\textwidth]{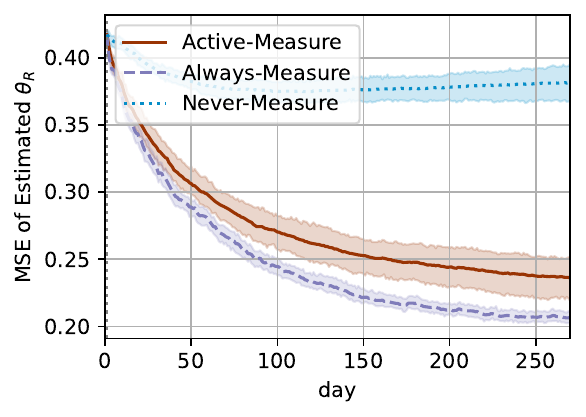}
        \caption{Medium +, zero $O_t$}
        \label{subfig:s12.mse}
    \end{subfigure}
    \caption{The MSE of $\widehat{\btheta}_t^{R(j)}$.}
    \label{fig:simulation.results.mse}
\end{figure*}

In addition, we report the mean squared error (MSE) of $\widehat{\btheta}_t^{R(j)}$, averaged over $J$ particles and 42 users, i.e.,
$
\frac{1}{42 J} \sum_{i = 1}^{42} \sum_{j = 1}^J \norm{\widehat{\btheta}_{t, i}^{R(j)} - \btheta_i^R}.
$
Figure~\ref{fig:simulation.results.mse} presents its mean and 95\% confidence interval over 50 replications.  
The figure shows that the MSE of active-measure is very close to that of always-measure, while the MSE of never-measure is significantly larger.  
This suggests that a small number of measurements is sufficient to obtain a near-optimal estimate of the transition function.  
Furthermore, when the emission $O_t$ is uninformative about the latent reward $R_t$, the MSE of never-measure increases substantially compared to when $O_t$ is informative.  

On average, active-measure takes about 38 minutes to complete one replication of the simulation for 42 users sequentially on a single CPU core of a cloud server, whereas always-measure and never-measure take about 34 minutes, and Dyna-ATMQ takes about 2 minutes.

\paragraph{Robustness of Active-Measure}
We conduct additional experiments using a misspecified transition model to evaluate the robustness of our proposed method.  
Specifically, we consider a general transition model from the current state $[E_{t-1}, R_{t-1}, C_{t}]$ to the next state $[M_{t, 2}, E_{t}, R_{t}, O_{t}]$ given the actions $[I_{t, 1}, A_{t, 2}]$.  
That is,
\begin{equation} \label{equ:general.transition.model}
\begin{split}
    C_{t, 2} &= \theta^C_{0} + \epsilon^C_{t, 2}, \\
    M_{t, 2} &= \theta^{M}_{0} + \theta^{M}_{1} E_{t-1} + \theta^{M}_{2} R_{t-1} + \theta^{M}_{3} C_{t, 2} 
    + A_{t, 2} (\theta^{M}_{4} + \theta^{M}_{5} E_{t-1} + \theta^{M}_{6} R_{t-1} + \theta^{M}_{7} C_{t, 2}) 
    + \epsilon^M_{t, 2}, \\
    E_{t} &= \theta^{E}_{0} + \theta^{E}_{1} E_{t-1} + \theta^{E}_{2} R_{t-1} + \theta^{E}_{3} C_{t, 2} 
    + A_{t, 2} (\theta^{E}_{4} + \theta^{E}_{5} E_{t-1} + \theta^{E}_{6} R_{t-1} + \theta^{E}_{7} C_{t, 2}) 
    + \theta^{E}_{I} I_{t,1} + \theta^{E}_{IE} I_{t,1} E_{t-1} 
    + \epsilon^{E}_{t}, \\
    R_{t} &= \theta^{R}_{0} + \theta^{R}_{1} E_{t-1} + \theta^{R}_{2} R_{t-1} + \theta^{R}_{3} C_{t, 2} 
    + A_{t, 2} (\theta^{R}_{4} + \theta^{R}_{5} E_{t-1} + \theta^{R}_{6} R_{t-1} + \theta^{R}_{7} C_{t, 2})
    + \epsilon^{R}_{t}, \\
    O_{t} &= \theta^{O}_{0} + \theta^{O}_{1} E_{t-1} + \theta^{O}_{2} R_{t-1} + \theta^{O}_{3} C_{t, 2} 
    + A_{t, 2} (\theta^{O}_{4} + \theta^{O}_{5} E_{t-1} + \theta^{O}_{6} R_{t-1} + \theta^{O}_{7} C_{t, 2})
    + \epsilon^{O}_{t},
\end{split}
\end{equation}
where $\theta^{E}_{I}$ and $\theta^{E}_{IE}$ are manually set to $-0.1$ and $0.01$, respectively.  
Since HeartSteps V2 includes $K = 5$ decision times per day, the general model was first fitted to HeartSteps with five actions and contexts.  
To construct the testbed with only one control action, we then aggregate the effects of the five contexts $C_{t, 1:5}$ and five actions $A_{t, 1:5}$ on $R_t$, $E_t$, and $O_t$, as described in Appendix~\ref{sec:testbed}.

Under this general testbed model, the working transition model described in Appendix~\ref{sec:posterior.parameters} for the proposed active-measure algorithm is misspecified.  
We compare it against the transition model specified as the true model in \eqref{equ:general.transition.model}.  
Only the emission model is misspecified as $O_{t} = \theta^{O}_{0} + \theta^{O}_{1} R_{t} + \epsilon^O_{t}$, since the emission must depend only on the current latent reward in the AOMDP framework.  
Furthermore, because $E_t$ is a function of the previous latent reward $R_{t - 1}$, it is also used to update the particle weight, similar to $M_{t, 2}$.  
In addition, since the mediational structure no longer exists, reward design is not applied in this setting.  
The model-free action selection algorithm remains the same as that described in Algorithm~\ref{alg:active.measure} for both transition models.  

\begin{figure*}
    \centering
    \includegraphics[width=0.32\textwidth]{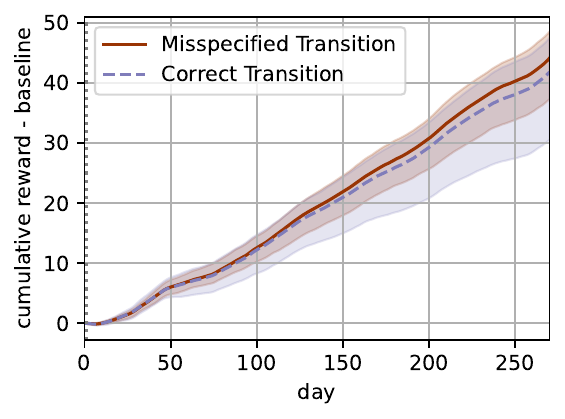}
    \caption{Comparing the average cumulative reward under misspecified and correctly specified transition models, subtracting the average cumulative rewards of the zero policy.}
    \label{fig:simulation.misspecified}
\end{figure*}

We compare the cumulative rewards—after subtracting the average cumulative rewards of the zero policy—in Figure~\ref{fig:simulation.misspecified}.  
We observe that the cumulative rewards are nearly identical under the misspecified and correctly specified transition models, with the misspecified model even exhibiting smaller variance.  
This demonstrates the advantage of using a parsimonious model in data-scarce settings: reducing the number of parameters may increase bias but can substantially reduce variance.

\section{Additional Related Work} \label{sec:additional.related.work}

The AOMDP extends the ACNO-MDP framework by allowing states $Z_t$ and emissions $O_t$ or $I_{t, 1} R_t$ to be observed between control and measure actions, thereby providing more accurate latent-state estimation.  
While $I_{t, 1}$ measures $U_{t}$ under our indexing, in ACNO-MDP the measurement action $I_{t-1}$ at time $t-1$ measures $U_{t}$.  
Our current indexing implicitly assumes that $U_{t}$ and $O_{t}$ occur before $I_{t, 1}$.  
The definition of $\bbO$, which depends only on the latent variable $U_{t}$, follows that in \citet{liu2022partially}.  

\citet{bellinger2021active} simultaneously chose the optimal control–measure action pair in tabular settings, but the learned policy always converged to non-measuring (see details in \citealp{krale2023act}).  
\citet{bellinger2022balancing} applied off-the-shelf deep RL algorithms to select the action pair in continuous settings based on the last measured state with a stale observation flag.  
\citet{nam2021reinforcement} proposed a heuristic for estimating latent states in continuous settings.  
They updated the transition function by maximizing the log-likelihood of the emission given the encoded history and action, and then drew particles from the estimated transition functions.  
However, the estimated unknown parameters and unobserved latent states are not necessarily drawn from their posterior distributions given the observed history.  
First, the transition parameters were estimated via maximum likelihood rather than by constructing a posterior.  
Moreover, the particle weights were not updated according to the SMC framework.  
Under SMC, the weights remain unchanged when the state is unmeasured and become proportional to the likelihood of the latent-state value given the history when the state is measured.  

Among the algorithms proposed for tabular settings, those introduced by \citet{krale2023act,krale2024robust} were heuristic methods that partially ignored future state uncertainty.  
\citet{avalos2024online} also distinguished between the states of the two actions and separated the two decision steps, but they assumed a pre-known model and focused on planning.  
In continuous-time RL, \citet{holt2023active} assumed noisy emissions and proposed an offline, continuous-time, model predictive control (MPC) planner.




\end{document}